\documentclass{article}

\usepackage[accepted]{dpaper2024}

\usepackage{xspace}
\usepackage{capt-of}
\def\eg{\emph{e.g.}\@\xspace}

 \def\vs{\emph{vs.}\@\xspace}

\definecolor{ER}{RGB}{229,173,213}
\definecolor{SOTA}{RGB}{118,184,133}
\definecolor{LwF}{RGB}{226,153,112}
\usepackage{booktabs}

\usepackage{multirow}

\usepackage{xcolor}
\usepackage{colortbl}
\usepackage{amsmath} 
\usepackage{amssymb}
\usepackage{graphicx}
\newcolumntype{s}{>{\columncolor{white!90!blue}} c}
\newcolumntype{g}{>{\columncolor{black!8}} c}
\newcolumntype{b}{>{\columncolor{black!3}} c}

\definecolor{cvprblue}{rgb}{0.21,0.49,0.74}
\usepackage[pagebackref,breaklinks,colorlinks,citecolor=cvprblue]{hyperref}

\dpapertitlerunning{Rethinking Continual Learning for Real-world Impact}

\begin{document}

\twocolumn[
\dpapertitle{Is Continual Learning Ready for Real-world Challenges?}

\begin{dpaperauthorlist}
\dpaperauthor{Theodora Kontogianni}{yyy}
\dpaperauthor{Yuanwen Yue}{yyy}
\dpaperauthor{Siyu Tang}{yyy}
\dpaperauthor{Konrad Schindler}{yyy}
\end{dpaperauthorlist}

\dpaperaffiliation{yyy}{ETH Zurich, Switzerland}

\dpapercorrespondingauthor{Theodora Kontogianni}{kontogianni.theodora@gmail.com}

\vskip 0.3in]

\printAffiliationsAndNotice{}  %

\begin{abstract}

 Despite continual learning’s long and well-established academic history  its  application in
real-world scenarios remains rather limited. This paper contends
that this gap is attributable to a misalignment between the
actual challenges of continual learning and the evaluation
protocols in use, rendering proposed solutions ineffective
for addressing the complexities of real-world setups. 
We validate our hypothesis and assess progress to date, using a new 3D semantic segmentation benchmark, OCL-3DSS. We investigate various continual learning schemes from the literature by utilizing more realistic protocols that necessitate online and continual learning for dynamic, real-world scenarios (\eg, in robotics and 3D vision applications). The outcomes are sobering: all considered methods perform poorly, significantly deviating from the upper bound of joint offline training. This raises questions about the applicability of existing methods in realistic settings. Our paper aims to initiate a paradigm shift, advocating for the adoption of continual learning methods through new experimental protocols that better emulate real-world conditions to facilitate breakthroughs in the field.

\end{abstract}    
\section{Introduction}
\label{sec:intro}
Continual learning\,(CL)\,\cite{ring1994continual} is one of the fundamental machine learning fields. Its methodology emerged as a solution for incorporating new knowledge in
existing models to adapt to new data distribution such as
new labels and tasks~\cite{ring1994continual}. Furthermore, in scenarios where re-training a model from scratch with the entire dataset is impractical or resource-intensive~(\eg foundation models), incremental updates on new data can make the process more efficient.

Despite continual learning's well-established academic history, spanning decades and tasks such as image classification~\cite{Aljundi_2018_ECCV,kirkpatrick2017overcoming,li2017learning,chaudhry2019continual,lopez2017gradient}, 2D semantic segmentation~\cite{michieli2019incremental,cermelli2020modeling,maracani2021recall}, and recently, 3D semantic segmentation~\cite{camuffo2023continual,yang2023geometry}, its practical application in real-world scenarios remains limited~\cite{gonzalez2020wrong}. This paper contends that this gap is attributable to a misalignment between the actual challenges of continual learning and the evaluation protocols in use, rendering proposed solutions ineffective for addressing the complexities of real-world setups.

This paper underscores the importance of recognizing the limitations of current evaluation protocols and advocates for exploring CL methods through new experimental protocols that mirror real-world setups. %
\textit{The results are surprising: All examined methods exhibit significant failure with near-zero intersection-over-union\,(IoU) even in medium-sized task sequences\,(20\,tasks) and fall considerably short of the upper bound achievable by joint offline training.}

While CL has traditionally focused on offline image classification~\cite{Aljundi_2018_ECCV,kirkpatrick2017overcoming,li2017learning,chaudhry2019continual,lopez2017gradient}, the demand for foundation models in autonomous agents and robotics, capable of dynamic adaptation in real-time 3D environments underscores 3D semantic segmentation as a more suitable task for benchmarking continual learning methods. 

Regardless of the task, CL methods  adhere to similar evaluation scenarios. These scenarios typically involve (i) multiple training epochs over the current task, allowing for the repetitive shuffling of data to facilitate task learning but not well-suited for streaming data. Additionally, (ii) reliance on large pre-trained models, while introducing only a few additional tasks in a CL scenario, often leading to trivial solutions. Lastly, (iii) a preference for a \textit{disjoint} scenario, where input data (pixels or 3D points) only from previous and current classes are included, an unrealistic assumption, particularly in the context of autonomous agents.

To address some of these limitations, there has been recently a growing interest in Online Continual Learning (OCL)~\cite{Ghunaim_2023_CVPR} that studies the problem of learning from an online stream of data and tasks. Such a setup is well-suited for dynamic environments where the learning process is ongoing such as in robotics but OCL methods primarily focus on image classification, as is common in most continual learning studies,  without delving into other tasks like semantic segmentation.

In a similar spirit, our setup features a single online data stream. At each time step, a small batch of data arrives, providing the model with an opportunity for learning before transitioning to the next incoming set of data~(\textit{single-pass}). Training unfolds over a prolonged sequence of tasks~(\textit{long task sequence}), where all classes are learned incrementally, starting immediately from the first task without any batch pre-training. Starting with a randomly initialized model, rather than one pre-trained on a set of tasks, may initially seem less realistic. However, it allows for the simulation of a more extended sequence of tasks in practice. Additionally, in contrast to previous methods emphasizing the \textit{disjoint} scenario, we prioritize the \textit{overlapped} scenario. Here, data points from future classes are also accessible, although without their corresponding ground truth.

 To study online continual learning in  3D semantic segmentation, we modified various well-known 3D semantic segmentation datasets,\,ScanNet\,\cite{Dai17CVPR}, S3DIS\,\cite{Armeni16CVPR} and SemanticKITTI\,\cite{behley2019semantickitti} to suit our online continual learning framework and used them to evaluate the performance of numerous widely-used continual learning approaches. Furthermore, inspired by recent approaches for continual learning in images we evaluated the use of a (i)\,vision-language model and a (ii)\,transformer decoder.

  CL suffers from the pervasive issue known as \textit{catastrophic forgetting}~\cite{french1999catastrophic, mccloskey1989catastrophic}.  This refers to the neural networks' tendency to abruptly and entirely erase previously acquired knowledge when learning new information.  %
  Online continual learning, which
assumes single-pass data, is strictly harder than offline, due
to the combined challenges of catastrophic forgetting and underfitting within a
single training epoch.

  Due to continual learning's long history, there are several works on continual learning from where we select a combination of seminal and state-of-the-art methods to fairly compare in our OCL-3DSS setup:\,MAS\,\cite{Aljundi_2018_ECCV} as a regularization-based approach, LwF\,\cite{li2017learning} as a distillation-based and ER~\cite{chaudhry2019continual} as a replay-based approach. Furthermore, we assessed the method proposed by Yang et al.~\yrcite{yang2023geometry}, which currently stands as the sole existing approach in continual learning for 3D semantic segmentation. The method primarily relies on the utilization of confident pseudolabels, as is common in many other CL methods. We additionally incorporated and evaluated language guidance and different backbone architectures.

  We observe that all methods exhibit comparable results when the number of tasks is small~($\leq5$), the common setup in existing benchmarks. However, their performance sharply declines to \underline{almost zero} after a few more tasks even for the recently popular methods with language guidance and transformer architecture. Notably, the replay-based method ER~\cite{chaudhry2019continual} stands out as the sole approach to maintain competitiveness in an OCL-3DSS scenario. However, with a mIoU of $42.1\%$ (20 tasks) and utilizing $33\%$ of the dataset in memory for replay, it still lags significantly behind the $66.4\%$ achieved on ScanNet by joint training.

Given the streaming nature of the task, we noticed that many classes are difficult to learn in an online continual learning setup without constant repetition of the class samples as in joint training. \textit{Positive backward} and \textit{positive forward transfer} has not been witnessed with the exception of positive backward transfer for the replay methods as it is expected when replaying previous samples. While replay-based methods may provide a practical solution, we contend that they do not fundamentally address the core aspects of continual learning and adaptation to new experiences. Detailed experiments supporting our assertion are presented in the subsequent sections of the paper.

We encourage future works to adopt our experimental protocol for assessing 2D and 3D semantic segmentation in a continual learning framework. Despite the challenging nature of our real-life setup, where most continual learning methods produce unfavorable results, we believe this is the way toward significant advancements in the field.

\section{Related Work}
\label{sec:relatedwork}
\vspace{-0.13cm}
\subsection{Offline Continual Learning}
The simplest and most common methods in continual learning have been focusing on two main approaches: replay-based and regularization-based methods. \\ 
\textbf{Replay-based methods} store input samples or representations of the current task while the model learns from them and later replays them while the model learns new tasks~\cite{lopez2017gradient,maracani2021recall,chaudhry2019continual}. Replay-based methods overcome catastrophic forgetting by simply re-training on the stored data of tasks already experienced so far. Some works store past experiences based on knowledge distillation~\cite{hinton2015distilling} especially when privacy is a concern but the simple storing of a set of raw samples from previous tasks~\cite{chaudhry2019continual,lopez2017gradient} works remarkably well. Replay-based methods are consistently the best-performing ones in continual learning. However, they are bound by the size of the replay buffer and the re-training time with every additional task. It could be argued that these approaches don't fundamentally tackle continual learning but rather represent an alternative manifestation of joint training. \\ 
\textbf{Regularization Methods.}
Several previous benchmarks on continual learning impose restrictions by prohibiting access to previously seen data. The rationale behind such constraints is often attributed to concerns related to storage space and privacy. The majority of these methods concentrate on addressing the issue of catastrophic forgetting through regularization, imposing penalties during training based on factors such as the importance of weights for specific tasks~\cite{Aljundi_2018_ECCV, kirkpatrick2017overcoming}. The objective is to identify crucial parameters for each task, restricting their alteration in subsequent tasks according to their performance. Alternatively, some approaches involve knowledge distillation from a prior model~\cite{li2017learning}. Notably, these methods operate without the need to store previously labeled examples. While they exhibit satisfactory performance in small datasets and straightforward classification tasks, their efficacy diminishes in challenging and complex scenarios~\cite{Ghunaim_2023_CVPR}, as corroborated by our experiments.\\
\textbf{Key Issues.} %
The majority of CL evaluations are conducted in small-scale datasets on classification tasks.  Furthermore, traditional continual learning methods start with a pre-trained model on a large subset of the tasks and incrementally introduce a handful of new tasks~(Fig.~\ref{fig:teaser}, Appendix). They allow the model to be trained over multiple epochs on each task with repeated shuffling of data and where the data distribution of each task is stationary.

\subsection{Online Continual Learning} 
To alleviate these limitations, recently OCL methods started focusing on the cases where data arrive
one tiny batch at a time and require the model to learn from a single pass of data~\cite{Ghunaim_2023_CVPR}. \\
\textbf{Key Issues.} Image classification remains the sole focus of all these methods as in most CL studies. In contrast, we introduce online learning for 3D semantic segmentation, a more pragmatic setup for autonomous agent tasks.

\subsection{Continual Learning in Semantic Segmentation} Recently, increased attention has been devoted to continual learning for semantic segmentation in 2D images~\cite{michieli2019incremental,cermelli2020modeling}. The problem was first introduced in~\cite{michieli2019incremental}. Reguralization-based PLOP~\cite{cermelli2020modeling} alleviates forgetting of old knowledge by distilling multi-scale features. Replay-based RECALL~\cite{maracani2021recall} obtains more additional data via GAN or web. Recently~\cite{camuffo2023continual} has presented a first attempt at using CL for
semantic segmentation on LiDAR point clouds in outdoor scenes.\\
\textbf{Key Issues.} All approaches addressing CL in Semantic Segmentation share the same foundational assumptions as traditional offline CL methods. These involve initiating with a pre-trained model on a substantial subset of tasks and gradually incorporating a few new tasks. Training occurs over multiple epochs for each task, allowing the repeated shuffling of data. The assumption is that the data distribution for each task remains stationary and is drawn from an i.i.d. distribution. \textit{In contrast, our approach introduces CL in 3D Semantic Segmentation by (i) adopting a fully incremental online setting and (ii) deviating from the i.i.d. assumption, making it more suitable for tasks involving autonomous agents.}
\section{Problem Formulation}
\label{sec:cl_eval_review}

Before we introduce our Online Continual Learning for 3D Semantic Segmentation~(OCL-3DSS) setup~(Sec.~\ref{subsec:oc3dcss}), we present the setup of semantic segmentation in 3D point clouds~(Sec.~\ref{subsec:batch_semseg}) and  offline continual learning~(Sec.~\ref{subsec:offline_cl}).

\subsection{Standard 3D Semantic Segmentation}
\label{subsec:batch_semseg}

Let \(x \in \mathbb{R}^{N \times C_{in}}\) be a 3D point cloud consisting of \(N\) points and \(C_{in}\) input channels (typically \(C_{in}=3\) for XYZ coordinates or \(C_{in}=6\) for XYZ and RGB if color information is available). The segmentation map, denoted as \(y \in \mathbb{R}^{N \times |\mathcal{C}|}\), represents semantic classes within the point cloud, where \(\mathcal{C}\) is the set of semantic classes.

Given a training set, the objective of semantic segmentation is to learn a model \(f_{\theta}(x): X \rightarrow Y\) that maps the input space \(X\) to a set of class probability vectors \(X \rightarrow \mathbb{R}^{N \times |\mathcal{C}|}\). The final prediction for each 3D point is determined by \(y_{\text{pred}} = \underset{c \in \mathcal{C}}{\text{argmax}} \ f_{\theta}(x)[i,c]_{i=1}^{N}
\), where \(f_{\theta}(x)[i,c]\) represents the probability of 3D point \(i\) belonging to class \(c\).

\subsection{Offline Continual Learning}
\label{subsec:offline_cl}

In contrast to conventional training approaches, continual learning~\cite{ring1994continual} involves training a machine learning model on a sequential stream of data from various tasks. Instead of learning from the entire dataset \(T \subset X \times Y\) at once, learning occurs in multiple steps denoted as \(t = 1, \ldots, T\), where only a subset of the classes \(\mathcal{C}_t \subset \mathcal{C}\) is available at each step \(t\). This new set of classes \(\mathcal{C}_t \cap \mathcal{C}_{1\ldots t-1} = \emptyset\) is used as the ground truth for updating the model parameters. The model is expected to perform optimally not only on the current task \(t\) but also to predict and perform well on all classes learned up to that point,

A key challenge in continual learning is the occurrence of \textit{catastrophic forgetting}~\cite{french1999catastrophic, mccloskey1989catastrophic}. This phenomenon arises when a network, trained with data from a new distribution corresponding to task $t$, tends to \textit{forget} previously acquired knowledge, resulting in a substantial decline in performance for tasks $1,\dots ,t-1$. From a broader CL perspective, this challenge reflects the \textit{stability-plasticity} dilemma~\cite{grossberg1982does}, where stability entails preserving past knowledge, while plasticity involves the ability to quickly adapt to new information.
\subsection{Towards A Realistic Evaluation}
\label{subsec:oc3dcss}

Continual Semantic Segmentation (CSS) aims to train a model \(f_{\theta}\) over \(T\) steps, where at each step \(t\), the model encounters a dataset \(D^t = \{x^t, \tilde{y}^t\}\). Here, \(x^t\) represents the input image or point cloud, and \(\tilde{y}^t\) denotes the ground truth segmentation at time \(t \in [1, \ldots, T]\). The segmentation map at each step includes only the current classes \(\mathcal{C}_t\), with labels from previous steps \(\mathcal{C}_{1:t-1}\) and future steps \(\mathcal{C}_{t+1:T}\) collapsed into a \textit{background} class or ignored. In the common \textit{disjoint} setup during the current task \(t\), input pixels or points \(i\) corresponding to future tasks \(x_i^t\) with \(\tilde{y}_i^{t+1:T}\) are excluded. Nevertheless, the model at step \(t\) is expected to predict all classes encountered over time, denoted as \(\mathcal{C}_{1:t}\).
The semantic segmentation task at time \(t\) is formulated as: \(\theta_t = \underset{\theta_t}{\text{argmin}} \, \mathop{\mathbb{E}}_{(x^t, \tilde y^t)} \mathcal{L}(f_{\theta}(x^t), \tilde y^t)\)

\textbf{Online Continual 3D Semantic Segmentation.} To establish a foundation for subsequent discussions, we introduce a straightforward continual learning process for sequential fine-tuning on a stream of 3D semantic segmentation tasks $(1,2,\dots)$. Starting with a randomly initialized model $f_{\theta_0}$, we iteratively adapt to models $f_{\theta_t}$ by incorporating $\tilde y^t$ into the current model $f_{\theta_{t-1}}$.

In contrast to prior works, we define our continual learning scenario in a more challenging setup, introducing key differences from previous setups in continual semantic segmentation~(CSS): (i) The model, $f_{\theta_0}$, is initialized with random weights, implying the absence of old tasks, only previous ones. (ii) In $D^t = \{x^t, \tilde{y}^t\}$, where $t \in \{\{1\}, \ldots, \{T\}\}$ and $t$ is a singleton set containing the labels of a single class. (iii) The number of tasks, $T$, is notably longer. (iv) Each dataset $D^t$ is processed only once.
\vspace{-4mm}
\section{Proposed Benchmark}

\subsection{A Closer Look At Existing Protocols}
In this section, we reflect on the existing evaluation protocols~\cite{yang2023geometry} and observe that they operate under the common disjoint setting of 2D class incremental segmentation, where the incremental training includes only the old and current classes of the point cloud, excluding the future classes. We find that existing setups adopt the two-set paradigm, one for \(\mathcal{C}_{\text{base}}\) and one for \(\mathcal{C}_{\text{novel}}\) classes. \(\mathcal{C}_{\text{novel}}\) contains only 5, 3, and 1 new classes for both ScanNet~\cite{dai2017scannet} and S3DIS~\cite{Armeni16CVPR}.

To put these numbers into perspective, in the case of \(\mathcal{C}_{\text{novel}} = 1\) in ScanNet, a pre-trained model with joint training for \(\mathcal{C}_{\text{base}} = 19\) classes is \underline{updated with a single class}. Classes are also split into \(\mathcal{C}_{\text{base}}\) and \(\mathcal{C}_{\text{novel}}\) based only on the original class label order of the dataset (\(S_0\)) or alphabetically (\(S_1\)), without taking into consideration the difficulty of each class. Furthermore, 3D semantic segmentation has witnessed tremendous improvement based on sparse convolutional architectures~\cite{Choy19CVPR} and transformers~\cite{schult2022mask3d}, which have enhanced the joint training upper bound even further. Notably, these architectures have not been considered in existing protocols.

We understand that the existing protocols in continual learning for 3DSS reflect existing 2D CSS benchmarks however we propose a more realistic setup that can reflect the progress in continual learning more accurately. Existing 3D semantic segmentation datasets can be modified to our OCL-3DSS protocol. To that end  we leverage three popular public 3D semantic segmentation datasets: (i)~ScanNet~\cite{dai2017scannet}, (ii)~S3DIS~\cite{Armeni16CVPR} and 
(iii)~SemanticKITTI~\cite{behley2019semantickitti}) to validate our evaluation setup.
\subsection{Datasets} We modify the following datasets for online continual 3D semantic segmentation: \\
\textbf{ScanNet}~\cite{dai2017scannet} is one of the largest real-world 3D datasets on the task of 3D semantic segmentation. It contains 1210 training scenes and 312 scenes for validation with 20 semantic labels. For training and validation, we follow the standard split for 3D semantic segmentation. We split the training set into 20 tasks, each one of them representing the learning of one label. Therefore, even though each 3D scene remains unchanged we use only the ground truth labels of the current class. We would like to process each scene a single time in a true online setup however since ScanNet is a highly imbalanced dataset (see Fig.~\ref{fig:scannetv2_stats} in Appendix) some object classes are present only in a few scenes. To alleviate the online learning problem and disentangle the task difficulty from the data availability we re-use scenes of the underrepresented classes by sampling with replacement until we have the maximum of 1201 scenes for each class. However, please note that this does not even the number of points labeled for each task~(for example classes like \textit{wall} remain one of the most represented ones making the learning still easier). After learning each class \(t\), we evaluate all classes up to and including \(t\) on the test set. When we reach class 20, the evaluation is identical to that in joint training. This protocol stands in direct contrast to existing setups, such as \(15-5\), \(17-3\), and \(19-1\)~\cite{yang2023geometry}. Our setup can be denoted as \(1-1-1 \cdots 1 = 1 \times 20\). Notably, we not only incrementally add classes, but this approach also results in a much longer sequence of tasks.\\
\textbf{S3DIS}~\cite{Armeni16CVPR} contains point clouds of 272 rooms in 6 indoor areas with 13 semantic classes. We use the challenging Area 5 as test and the rest as training in an online continual learning setup as above. We use all classes except \textit{clutter}.\\
\textbf{SemanticKITTI}~\cite{behley2019semantickitti} is an outdoor dataset without RGB information (compared to ScanNet and S3DIS). 
It is a very big dataset so to align with the online setup we selected sequence 0 for training and sequence 8 for testing.
We use 9 classes present in those scenes for our online continual learning setup: \textit{car, road, sidewalk, building, fence, vegetation, trunk, terrain} and \textit{pole}. 

\subsection{Task Difficulty Scenarios}
\label{sec:task_diff}
Since the order of arrival of each task could influence the performance we organize three different learning scenarios: \textit{1.~standard}, \textit{2.~difficult$\rightarrow$easy}, \textit{3.~easy$\rightarrow$difficult}. The \textit{standard} refers to the existing order from the dataset creators, difficult$\rightarrow$easy
introduces tasks from the rarest to the most common class based on the number of 3D points of each class and easy$\rightarrow$difficult uses the inverse order.
\subsection{Decoder Architectures}
We explore two decoder architectures. One is a standard convolutional head that predicts semantic logits. In addition, we adapt the recent Transformer decoders~\cite{strudel2021segmenter,cheng2022masked} for image semantic segmentation to our online continual 3D semantic segmentation. We show our findings are independent from specific decoder architectures.

\section{Class  Supervision Using Language Models}
\label{sec:clip}
Recently, prompt tuning has surfaced as an alternative to rehearsal buffers in continual learning for 2D semantic segmentation. Approaches in this domain~\cite{wang2022learning,wang2022dualprompt,khan2023introducing} leverage a pre-trained vision encoder and employ prompt learning to facilitate the continuous learning of tasks, eschewing the need to replay samples from preceding tasks. 

Integrating language into continual learning for 3D semantic segmentation presents challenges due to the scarcity of images typically found in such datasets. In the absence of readily available images, our reliance is solely on text prompts for guidance and instruction.

For a given training sample \((x, y_t)\) consisting of a point cloud \(x\) with point labels \(y_t\), we express the class name for class \(t\) in language as the following prompt:

\begin{center}
     ``A photo of \color{purple}{class name}\color{black}''
\end{center}

The prompt is input to a pre-trained text encoder to extract the feature corresponding to the output token, representing \(E^t_{lang} \in \mathbb{R}^C\), the language representation of class \(t\) with embedding dimension \(C\).
 
We aim to constrain the feature encodings \(E_l^t\) of class \(t\) in the model at layer \(l\) to be close to the language representation of the same class. We introduce language guidance through this feature using cosine similarity:

\begin{equation}
    S(E^t_{\text{lang}}, E_l^{t}) = \dfrac{E^t_{\text{lang}} \cdot E_l^{t}}{\|E^t_{\text{lang}}\| \cdot \|E_l^{t}\|}
\end{equation}

\begin{figure}[t!]
    \centering
    \begin{minipage}{0.47\columnwidth}
        \centering
        \includegraphics[width=\linewidth]{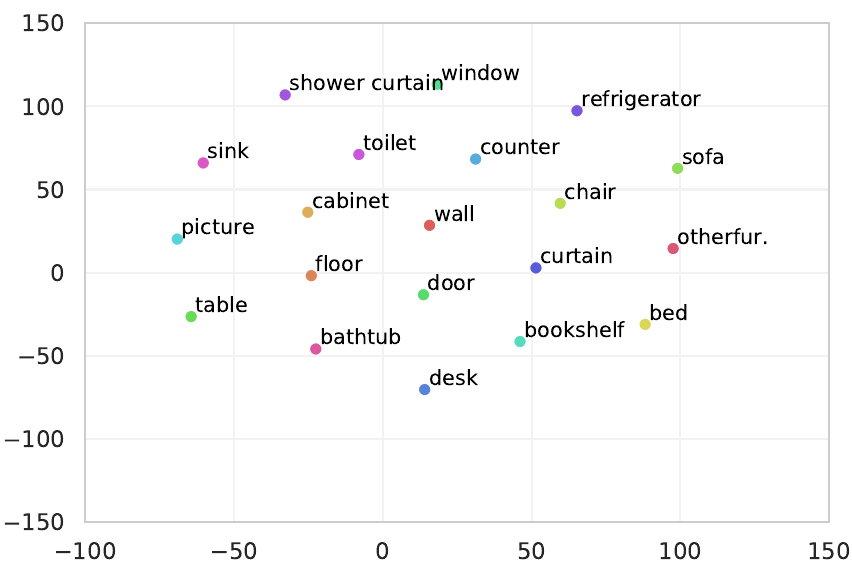}
    \caption{\textbf{T-SNE of CLIP features on ScanNet~\cite{Dai17CVPR} classes,}  evenly distributed in 2D space.}
    \label{fig:tsne}
    \end{minipage}%
    \hfill
    \begin{minipage}{0.47\columnwidth}
    \centering
    \includegraphics[width=\linewidth]{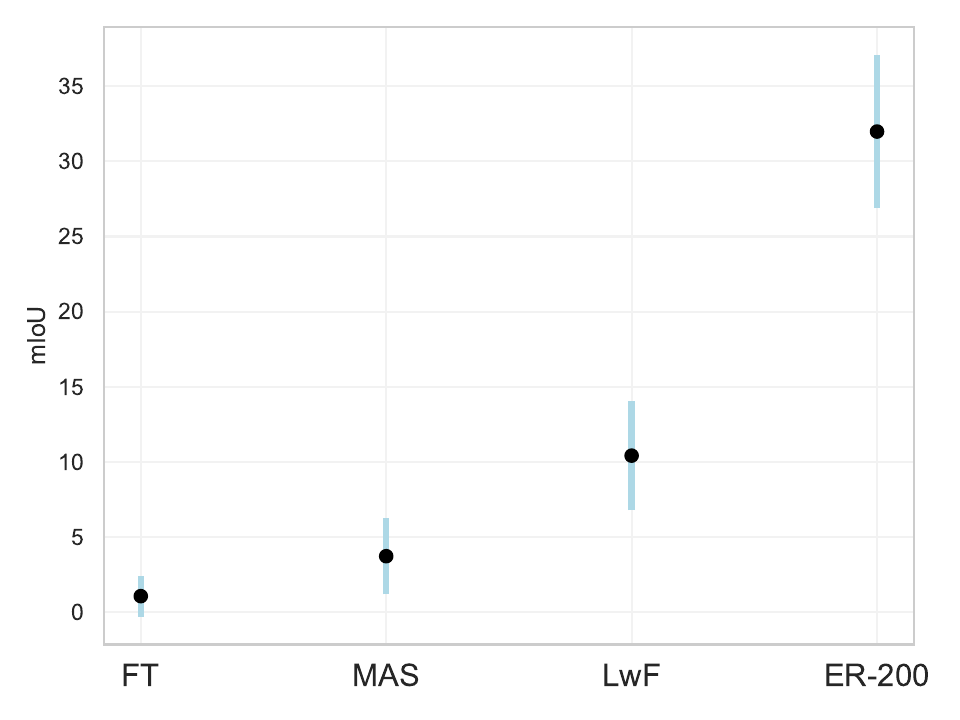}~
    \caption{Performance variance on the task ordering in ScanNet.}
    \label{fig:errorbar}
    
    \label{fig:ed}
    \end{minipage}
    \end{figure}

We define \(S(E^t_{\text{lang}}, E_l^{t'})\) similarly. We optimize the cosine loss to incentivize the model to align the per 3D point features of the network close to the language representation of their respective class \(E^t_{\text{lang}}\). Simultaneously, we mitigate the forgetting of previous tasks by keeping the features corresponding to all other classes \(t'\) far away from \(E_{\text{lang}}^t\). This leverages the fact that the text features are already separable (Fig.~\ref{fig:tsne}), thus alleviating forgetting.

Given the task \(t\), the model only has access to the current task labels and no access to labels from other tasks. Therefore, we optimize the following triplet loss based on the cosine similarity described above:
\begin{equation}
    L(A, P, N) = -S(A, P) + S(A, N) + \text{margin}
\end{equation}
\vspace{-1ex}
The total optimization is defined as:
\begin{equation}
    J^t = \sum_{i=1}^M L(A^t, P^i, N^i)
\end{equation}
Here, we define the anchor point \(A^t\) as the feature encoding \(E^t_{\text{lang}}\), \(P = E^t_{l}\) as the feature vector of points belonging to class \(t\), and \(N = E^{t'}_{l}\) as the feature vector of points not belonging to class \(t\). We sample an equal number \(M\) of points \(i\) for both the negative and positive points.

During inference, we assign a label to each point based on the maximum similarity to the language features. This way we require no adaptive decoder and we can add as many classes as needed.

\vspace{10mm}

\section{Experiments}

\begin{figure}[t!]
    \centering
    \begin{minipage}{0.49\columnwidth}
        \centering
                \vspace{-5pt}
        \includegraphics[width=0.98\linewidth]{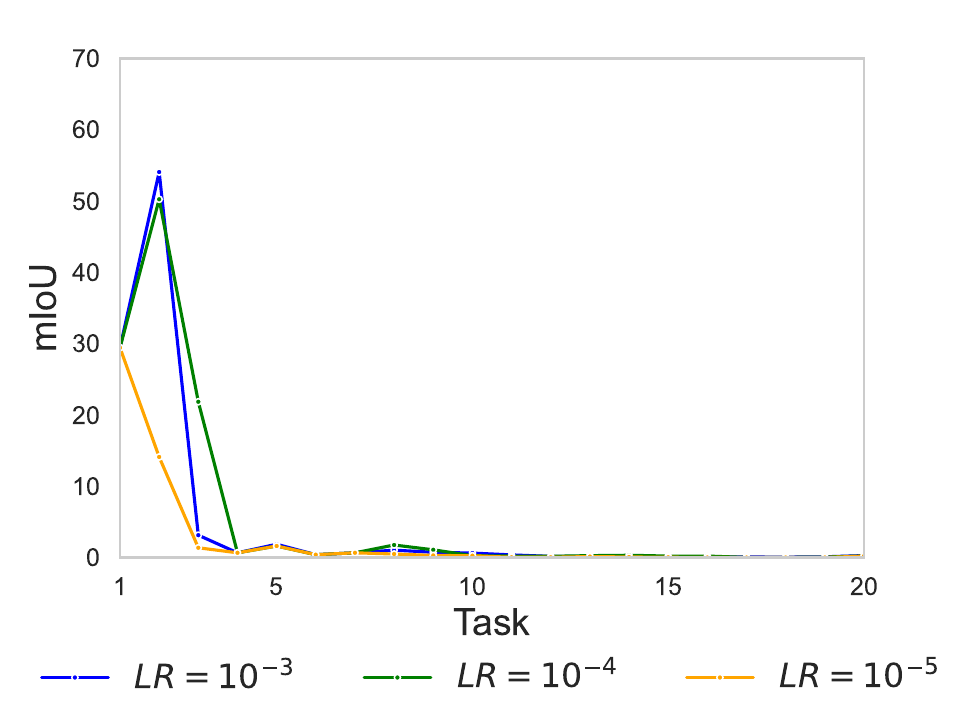}
    \vspace{0ex}
    \caption{Learning rate in FT has minimal influence.}
    \label{fig:lr}
    \end{minipage}
    \hfill
    \begin{minipage}{0.49\columnwidth}
\includegraphics[width=0.98\linewidth]{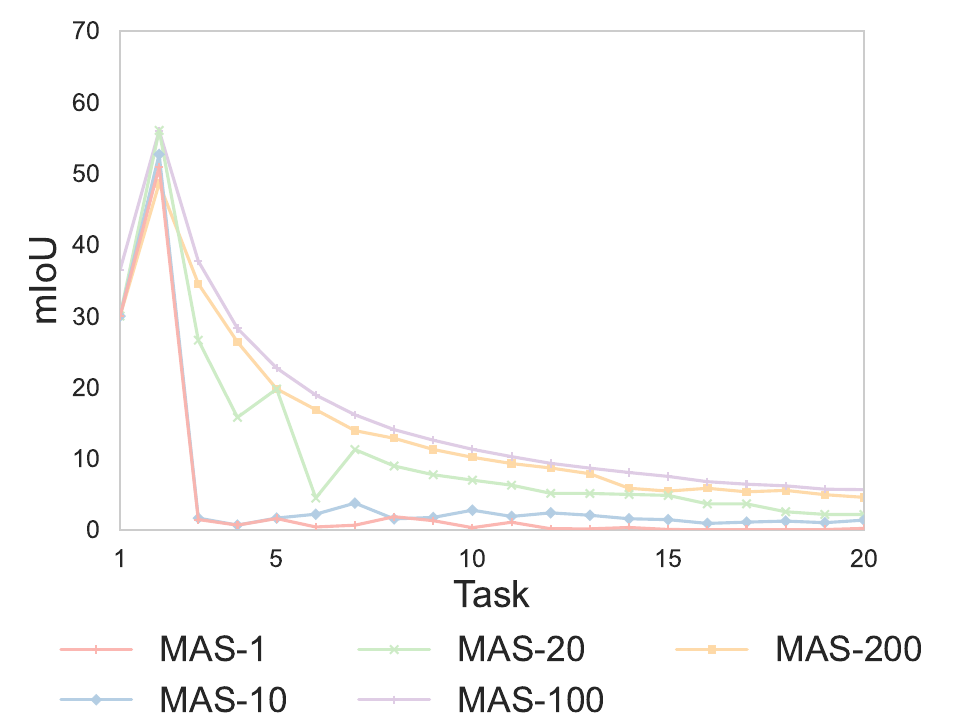}
    \vspace{-1ex}
    \caption{MAS weight does not significantly impact forgetting.}
    \label{fig:mas_weight}
    \end{minipage}%
    \vspace{1ex}
    \end{figure}

\textbf{Existing Approaches and Baselines.} We consider the following state-of-the-art approaches for
our empirical studies: MAS~\cite{Aljundi_2018_ECCV}, LwF~\cite{li2017learning} and ER~\cite{chaudhry2019continual}. We choose them as canonical
examples for regularization-based, distillation-based and memory replay approaches, discussed in Section~\ref{sec:relatedwork}. 
LwF regularizes the output of the network in task $t$ based on the outputs of the network of task $t-1$ by always saving the model parameters of the previous task. MAS penalizes the change of the important parameters for the previous tasks basing their importance on the sensitivity of the predicted output to the change in the parameter. ER as in \cite{Ghunaim_2023_CVPR} performs one gradient step on a batch, half of which is randomly sampled from the memory. We also evaluate
Yang et al.~\yrcite{yang2023geometry} as the only baseline in continual learning for 3D semantic segmentation that combines pseudo-labels created by the model from previous tasks with a distillation loss.

Finally, we also compare with
a reference model learned in a traditional semantic segmentation joint training (JT) without continual learning,
which may constitute an upper bound. Our lower bound consists of fine-tuning with new task data~(FT) without taking any measures to alleviate catastrophic forgetting.\\
\textbf{Metrics}. We report mean \textit{Intersection over Union}~(mIoU) both after the final incremental step and at the intermediate stages. Notably, while many methods demonstrate satisfactory performance for up to three additional tasks, there is a tendency to completely forget earlier tasks in longer sequences. We additionally report: 
\begin{equation*}
    \textit{forgetting}_t = \dfrac{\text{IoU}_t - \text{IoU}_t^T}{\text{IoU}_t} \times 100\%
\end{equation*}
where $\text{IoU}_t$ is the current task $t$ performance at step $t$ and $\text{IoU}_t^T$ is the performance of task $t$ after all the steps. The smaller the value, the better. Negative values mean improved learning in future tasks.\\
\textbf{Implementation Details.} During each training step $t$, a batch consists of 10 3D scenes. In the case of ER, 5 scenes are randomly sampled from memory. This process repeats until all scenes for each task are utilized. To handle the long-tail nature of 3D semantic segmentation, where some classes appear only in a limited number of scenes, we address this by resampling scenes until the count aligns with the maximum available scenes of the most prevalent task.\\
\textbf{Backbones.} We use the MinkowskiEngine\,(ME)~\cite{Choy19CVPR} with a $5~cm$ voxel size as a backbone that we train from scratch starting on task 1.  We use SGD with a fixed common learning rate for all tasks. Besides the standard ME decoder, we also modify a Transformer decoder~\cite{strudel2021segmenter,cheng2022masked} to our task. For the language guidance LG-CLIP, we use CLIP~\cite{radford2021learning}.\\
\textbf{Supervision}.
All methods are fully supervised with all the 3D points with the label of the current class. 
When language guidance is used supervision requires only 10 points per class per scene and we follow that scenario in all our table entries that report CLIP. The supervision of the other methods is at the level of at least hundreds of points per scene (depending on the class popularity, even thousands).

\section{Analysis and Discussion}
\begin{figure}[t]
    \centering
\includegraphics[width=0.8\linewidth]{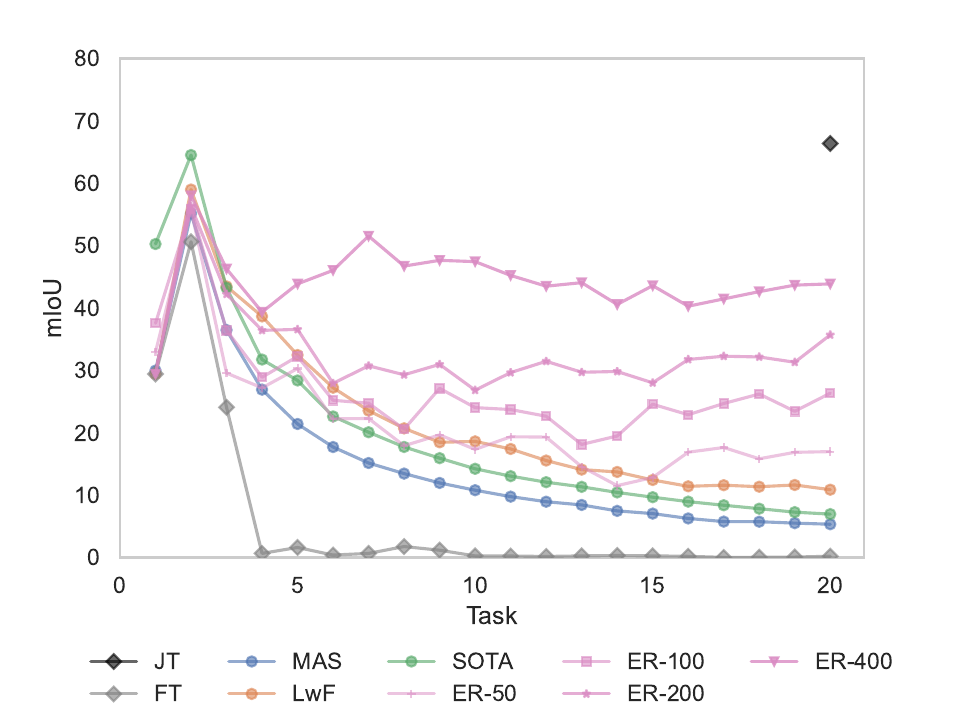}
     \vspace{-10px}
    \caption{\textbf{Overall performance} of CL methods on our OCL-3DSS setup (ScanNet-v2). All CL methods except ER converge to almost zero by the $20_{th}$ task. Despite the sizable memory buffer, even ER is still far from JT.  ER is more of a practical solution.}
    \label{fig:total}
\end{figure}

\begin{table}[h!]

    \caption{\textbf{OCL-3DSS} on ScanNet-val~\cite{dai2017scannet} after 3, 5, 10 and 20~(final) tasks. Augmenting the point clouds did not have an influence on other methods besides the FT. Using augmentation for training in cases where the i.i.d assumption does not hold (eg. multiple scans of the same scene) might be comparable to MAS and LwF for few tasks~($\leq$5).}
    \label{tab:scannet_miou_aug}
    \vskip 0.1in
\setlength{\tabcolsep}{8pt}
    
\centering
    \resizebox{0.7\columnwidth}{!}{%

    \begin{tabular}{lp{7mm}p{6mm}p{6mm}p{6mm}}
    \toprule
Method  &\multicolumn{4}{c}{\textbf{mIoU after \textit{t} tasks}}\\
& \textbf{3} & \textbf{5}  & \textbf{10}  & \textbf{20}\\
\cmidrule{1-5}
\tt{FT}~(\textit{lower bound}) & 24.9 & 3.9 & 2.1 & 0.4\\
\tt{FT} + Aug.  & \textbf{27.3} & \textbf{17.4} & \textbf{9.1} & \textbf{4.1}\\
         \bottomrule
    \end{tabular}}

    \caption{\textbf{OCL-3DSS} on ScanNet-val~\cite{dai2017scannet} after 3, 5, 10 and 20~(final) tasks with \textit{dice loss} instead of cross entropy. Only  ER~\cite{chaudhry2019continual} improved. The other methods had comparable results~(see Tab.~\ref{tab:scannet_miou_wdice_appendix} in Appendix).}
    \label{tab:scannet_miou_wdice}
    \vskip 0.1in
\centering
    \resizebox{0.87\columnwidth}{!}{%

    \begin{tabular}{lcccc}
    \toprule

Method &\multicolumn{4}{c}{\textbf{mIoU after \textit{t} tasks}}\\
& \textbf{3} & \textbf{5}  & \textbf{10}  & \textbf{20}\\
\cmidrule{1-5}

\tt{ER-200}\cite{chaudhry2019continual}   & 42.2 & 36.6 & 26.8 & 35.7\\

\tt{ER-200\cite{chaudhry2019continual}}+  DiceLoss & \textbf{47.1} & \textbf{41.7} & \textbf{43.8} &\textbf{36.1}\\

\bottomrule
    \end{tabular}}
\end{table}

\begin{table}
\vspace{-5px}
\caption{\textbf{OCL-3DSS} on ScanNet-val~\cite{dai2017scannet} after 3, 5, 10 and 20~(final) tasks with \underline{limited} use of memory. Just a single stored example from each class performs almost as well as 50. With the additional language guidance, it can surpass the 50 examples in the 5 and 10 classes. The size of the memory matters actually at the 20 classes. So storing several examples per task is important for large task sequences but few examples need to be stored for a moderate number of tasks.
}
\label{tab:lowmem}
\vskip 0.1in
\setlength{\tabcolsep}{25pt}
    
\centering
    \resizebox{\columnwidth}{!}{%

    \begin{tabular}{lp{8mm}p{6mm}p{6mm}p{6mm}}
    \toprule
Method  &\multicolumn{4}{c}{\textbf{mIoU after \textit{t} tasks}}\\
& \textbf{3} & \textbf{5}  & \textbf{10}  & \textbf{20}\\
\cmidrule{1-5}
\tt{ER-50}\cite{chaudhry2019continual} & 29.6& 30.3 & 17.4 & 17.0\\
\midrule
\tt{ER-2}\cite{chaudhry2019continual}  & 37.0& 29.9 & 18.3 & 14.1\\
\tt{ER-1}\cite{chaudhry2019continual}  & 31.7& 24.8 & 18.5 & 8.8\\
\tt{ER-1+LG-Clip}  & 30.5 & 34.1 & 20.7 & 8.1\\
         \bottomrule
    \end{tabular}}

\end{table}

\begin{table}[t]
\caption{\textbf{OCL-3DSS} on ScanNet-val~\cite{dai2017scannet} after 3, 5, 10 and 20~(final) tasks. Best method is ER-400. Most methods offer comparable results up to the $5_{th}$ task but only replay methods maintain performance after. LG-CLIP is weakly supervised with just 10 3D points per class.}
\label{tab:scannet_miou}
\vskip 0.1in
    \centering
    \resizebox{\columnwidth}{!}{%
\setlength{\tabcolsep}{8pt}

    \begin{tabular}{lrccccg}
    \toprule
Method & \textbf{Mem. Size}& \textbf{Mem./Data}  &\multicolumn{4}{c}{\textbf{mIoU after \textit{t} tasks}}\\
&  $slots\cdot |C|$ &(percent) & \textbf{3} & \textbf{5}  & \textbf{10}  & \textbf{20}\\
\cmidrule{1-7}

\tt{FT(lower bound)} &  - & - & 24.9 & 3.9 & 2.1 & 0.4\\
\cmidrule{1-7}

\tt{MAS\cite{Aljundi_2018_ECCV}} & - &-   & 33.7 & 22.6 & 10.8 & 4.9 \\
\tt{LwF\cite{li2017learning}}  &- & -&41.2 & 34.1& 17.4 & 7.8\\
  \cmidrule{1-7}
\tt{ER-50}\cite{chaudhry2019continual} & 50  $\cdot$ 20 & 4.1& 29.6& 30.3 & 17.4 & 17.0\\
     \tt{ER-200}\cite{chaudhry2019continual} & 200  $\cdot$ 20 & 16.4  & 42.2 & 36.6 & 26.8 & 35.7\\
\tt{ER-400}\cite{chaudhry2019continual} & 400  $\cdot$ 20 & 32.8  & \underline{43.6} & \underline{36.4} & \underline{39.0} & \underline{42.1}\\
\cmidrule{1-7}
\tt{Yang et al.}\yrcite{yang2023geometry} &&&40.7& 24.3 &12.1 &7.0\\
\cmidrule{1-7}

\tt{LG-Clip-weak sup.}  & - & - & 37.5   & 18.5  
&  10.4 & 3.2\\
\cmidrule{1-7}

\tt{JT}(upper bound) & - & -& -& - & -& \textbf{66.4} \\
\bottomrule

\end{tabular}}

\end{table}

\begin{table}[t!]
\vspace{-5px}
\caption{\textbf{OCL in 3D semantic segmentation} on S3DIS-A5~\cite{Armeni16CVPR} after 3, 6 and 12~(final) tasks. }
    \label{tab:s3dis_miou}
    \vskip 0.1in
    \centering
    \resizebox{\columnwidth}{!}{%
\setlength{\tabcolsep}{8pt}

    \begin{tabular}{lrcccg}
    \toprule

Method&     \textbf{Mem. Size}& \textbf{Mem./Data}  &\multicolumn{3}{c}{\textbf{mIoU after \textit{t} tasks}}\\
&   $slots\cdot |C|$ & (percent) &\textbf{3}   & \textbf{6}  & \textbf{12}\\
\cmidrule{1-6}

\tt{FT}(lower bound) &  - & - & 25.1 & 16.3 & 9.1 \\
\cmidrule{1-6}

\tt{MAS}\cite{Aljundi_2018_ECCV} & - &-   & 29.5 & 14.7 & 8.6  \\
\tt{LwF}\cite{li2017learning}  &- & -&30.8 & 8.3 & 6.4  \\
\cmidrule{1-6}

\tt{ER-20}\cite{chaudhry2019continual} & 20  $\cdot$ 12 & 9.8 & 33.2& 17.4 & 8.6 \\
\tt{ER-50}\cite{chaudhry2019continual} & 50  $\cdot$ 12& 24.5   & 33.7 & 18.3 & 13.7  \\
\tt{ER-100}\cite{chaudhry2019continual} & 100  $\cdot$ 12 & 49.0  & \underline{34.3} & \underline{19.4} & \underline{16.5}\\
\cmidrule{1-6}

\tt{LG-Clip}-weak sup.  & - &-  & 34.2 & 11.6 & 4.4\\
\cmidrule{1-6}

\tt{JT}(upper bound) & - & -& -& - &  \textbf{65.4} \\
 \bottomrule

    \end{tabular}}
    
\end{table}

\begin{table}
\caption{\textbf{OCL-3DSS} on  SemanticKITTI~\cite{behley2019semantickitti} after 3, 5 and 9\,(final) tasks. }
    \label{tab:kitti_iou}
    \centering
    \vskip 0.1in
    \resizebox{\columnwidth}{!}{%
\setlength{\tabcolsep}{8pt}

    \begin{tabular}{lrcccg}
    \toprule

        Method  & \textbf{Mem. Size}& \textbf{Mem./Data}  &\multicolumn{3}{c}{\textbf{mIoU after \textit{t} tasks}}\\
          &   $slots\cdot |C|$ & (percent) &\textbf{3} & \textbf{5}  & \textbf{9}\\
 \cmidrule{1-6}

\tt{FT}(lower bound) &  - & - & 6.4 & 4.9 & 3.4\\
\cmidrule{1-6}

\tt{MAS}\cite{Aljundi_2018_ECCV} & - &-   & 7.4 & 3.3 & 7.1 \\
\tt{LwF}\cite{li2017learning}  &- & -&41.1 & 44.5 & 49.1 \\
      \cmidrule{1-6}

\tt{ER-200}\cite{chaudhry2019continual} & 200  $\cdot$ 9 & 4.4 & 29.0& 33.5 & 48.5 \\
  \tt{ER-500}\cite{chaudhry2019continual} & 500  $\cdot$ 9& 11.0   & \underline{29.0} & \underline{35.2} & \underline{52.5} \\
         \tt{ER-1000}\cite{chaudhry2019continual} & 1000  $\cdot$ 9 & 22.0  & {31.0} & 34.4 & {52.3} \\
\cmidrule{1-6}

\tt{LG-Clip-weak sup.}  & - & - &  3.2&  8.1 & 2.2\\
\cmidrule{1-6}

\tt{JT}(upper bound) & - & -& -& - & \textbf{67.4} \\
 \bottomrule

    \end{tabular}}
    
\end{table}

\begin{table*}[t]
\caption{\textbf{\textit{Forgetting} per class on ScanNet-val~\cite{Dai17CVPR}}. All methods except ER converge to almost zero IoU.  Methods like LwF~\cite{li2017learning} and Yang et al.~\yrcite{yang2023geometry} try to alleviate it but they just remember the populous classes like \textit{wall} and \textit{floor} and fail to learn new ones. With - we represent \textbf{underfitting}\,(zero IoU while learning that class). Negative scores are good $\rightarrow$ mIoU is increasing with the progression of tasks. Possible only with replay or pseudolabels.}\label{tab:perstep-C2F}
 \label{tab:forgets}
 \vskip 0.1in
\centering
\resizebox{0.93\linewidth}{!}{%
\setlength{\tabcolsep}{10pt}
\begin{tabular}{l|cccccccccccccccccccc}
\toprule
\textit{Forgetting}$\downarrow$\\Method & \rotatebox{90}{wall $\downarrow$} & \rotatebox{90}{floor  $\downarrow$} & \rotatebox{90}{cabinet $\downarrow$} & \rotatebox{90}{bed $\downarrow$} & \multicolumn{1}{c}{\rotatebox{90}{chair $\downarrow$}} & \rotatebox{90}{sofa} & \rotatebox{90}{table $\downarrow$} & \multicolumn{1}{c}{\rotatebox{90}{door}} & \rotatebox{90}{window $\downarrow$} & \rotatebox{90}{bookshelf$\uparrow$} & \rotatebox{90}{picture$\uparrow$} & \multicolumn{1}{c}{\rotatebox{90}{counter $\downarrow$}} & \rotatebox{90}{desk $\downarrow$} & \multicolumn{1}{c}{\rotatebox{90}{curtain $\downarrow$}} & \rotatebox{90}{fridge $\downarrow$} & \multicolumn{1}{c}{\rotatebox{90}{sh.curtain $\downarrow$}} & \rotatebox{90}{toilet $\downarrow$} & \rotatebox{90}{sink $\downarrow$} & \rotatebox{90}{bathtub $\downarrow$} & \rotatebox{90}{\textbf{mIoU $\uparrow$} }   \\
 \midrule

\tt{FT} & 100 & 100 & 100 & 99 & 99 & 95 & 100 & 98 & 100 &100 & 100 &100  & 100 & 100 & 100 & 100 & 100 & 100 & 100     & 0.4    \\
\tt{MAS}\cite{Aljundi_2018_ECCV} & -52 & 3 & 72 & 11 & - & 52 & - & - & - &- & - &-  & - & - & - & - & - & -   & -    & 4.9    \\
\tt{LwF}\cite{li2017learning} &  -32& 3 & - & -  & - &  & - & - & - &- & - &-  & - & - & - & - & - & - & -    &    7.8  \\
\tt{ER-50}\cite{chaudhry2019continual} & -58 & -43 & 100 & -1467 &  -116 & -282 & -18 & 67 & -75 &-367 & 100 &-67  & -47 & -225 & 100 & -200 & -533 & 100 & -140   &  17.0   \\
\midrule
\tt{Yang et al.\yrcite{yang2023geometry}}&  -0.3& -4 &-  &-  & -0.1 & - & - & - & - &- & - &-  & - & - & - & - & - & -  & -  &  7.0 \\
\midrule
\tt{LG-CLIP-weak sup.}&  30&	52	&95 &	79 &	96 &	96	 &	-1 &  	92 &	95 &	62 &	5 &	-884	&-142 & -115	&-9  &	46	 &-200 &	42 &	-1     & 3.2\\
\bottomrule

    \end{tabular}}
            
\end{table*}

\textbf{The challenge: Differences between JT and FT.} We first study the impact of incremental fine-tuning, in a relatively long sequence of tasks 20, 12, and 9 for ScanNet, S3DIS and SemanticKITTI respectively. There is a big discrepancy between the Joint Training (\texttt{{JT}}) of a set of tasks and Finetuning~(\texttt{FT}) in each task sequentially as shown in Tables~(\ref{tab:scannet_miou}, \ref{tab:s3dis_miou}, \ref{tab:kitti_iou}) and Fig.~\ref{fig:total}. For example, as it can be seen in Tab.~\ref{tab:scannet_miou},
JT on the 20 classes of the ScanNet dataset with a batch size of 10 for 25 epochs returns an mIoU of 66.4 in the validation set. This can be considered the upper bound for the task on such a backbone. To understand the scale of the catastrophic forgetting issue the sequential training of tasks without any countermeasures to mitigate the catastrophic forgetting yields 0.4 mIoU (Fig.~\ref{fig:total}). Besides \textit{catastrophic forgetting}, online learning of tasks sequentially 
encounters an additional challenge: learning with limited data~(Tab.\,\ref{tab:forgets}).\\
\textbf{Is OCL purely an optimization problem?}
Adjusting the learning rate\,(Fig.\,\ref{fig:lr}) does not alleviate forgetting nor does augmentation\,(Tab.\,\ref{tab:scannet_miou_aug}). Dice loss\,(Tab.\,\ref{tab:scannet_miou_wdice})  alleviates the class imbalance and facilitates learning new classes only in ER. \\
\textbf{Are regularization methods a solution?} We have evaluated one of the most popular regularization-based methods MAS~\cite{Aljundi_2018_ECCV}. MAS has shown good results in classification tasks when using a pre-trained network on ImageNet where new tasks are added~(a popular setup in offline CL on classification tasks). However, in our challenging setup of incremental learning from scratch the method performs poorly~(Tables~\ref{tab:scannet_miou},\ref{tab:s3dis_miou},\ref{tab:kitti_iou} and Fig.~\ref{fig:mas_weight}). Different weights for the MAS loss compared to the segmentation loss do not seem to have a significant effect on the final outcome~(Fig.~\ref{fig:mas_weight}). Large weight values just reduce the plasticity. Contrary to~\cite{Aljundi_2018_ECCV}, in our case, LwF~\cite{li2017learning} outperforms MAS  showing that in our more challenging setup, the existing takeaways from the CL field of image classification do not necessarily hold. It is very important to point out that both methods experience catastrophic forgetting almost after the $3_{rd}$ task that cannot be mitigated both in ScanNet and S3DIS. In SemanticKITTI~(Tab.~\ref{tab:kitti_iou}) LwF seems to perform very well.\\
\textbf{Is it possible to CL without memory?} As shown in Tables~\ref{tab:scannet_miou},\ref{tab:s3dis_miou},\ref{tab:kitti_iou} the best performing method is ER~\cite{chaudhry2019continual}. It is a simple method where half of the training batch consists of samples of previous tasks randomly selected. In our case, we store the raw labels of the 3D points of previous tasks. As anticipated, the size of the memory is a relevant factor, at least up to a certain threshold, especially as the number of tasks increases. It's crucial to highlight that despite storing nearly one-third of our dataset, the performance at the conclusion of 20 tasks reaches, at best, approximately 60\% of the upper limit for ScanNet~\cite{Dai17CVPR} (Tab.~\ref{tab:scannet_miou}). Comparable findings are observed in  S3DIS~\cite{Armeni16CVPR}~(Tab.~\ref{tab:s3dis_miou}) and SemanticKITTI~\cite{behley2019semantickitti}~(Tab.~\ref{tab:kitti_iou}).
The second-best approach is LwF~\cite{li2017learning}, which enforces regularization on the new model predictions, compelling them to align closely with the predictions of the model from the previous task. Following closely in performance is the method proposed by Yang et al.~\yrcite{yang2023geometry}, which derives much of its effectiveness from utilizing pseudolabels generated from confident predictions in previous tasks. However, if objects from the previous class are absent in the scene, the method struggles, resulting in the retention of only ubiquitous classes like walls and floors that are present in every scene (Tab.~\ref{tab:forgets}). It is noteworthy to reiterate that both methods experience a significant performance decline after the 5th task, reinforcing our assertion that online CL and CL, in general, necessitate much longer sequences of tasks for a comprehensive evaluation. SemanticKITTI stands out as an exception, given that the majority of its classes are present in every scene.
\begin{figure}
    \centering
\includegraphics[width=0.5\linewidth]{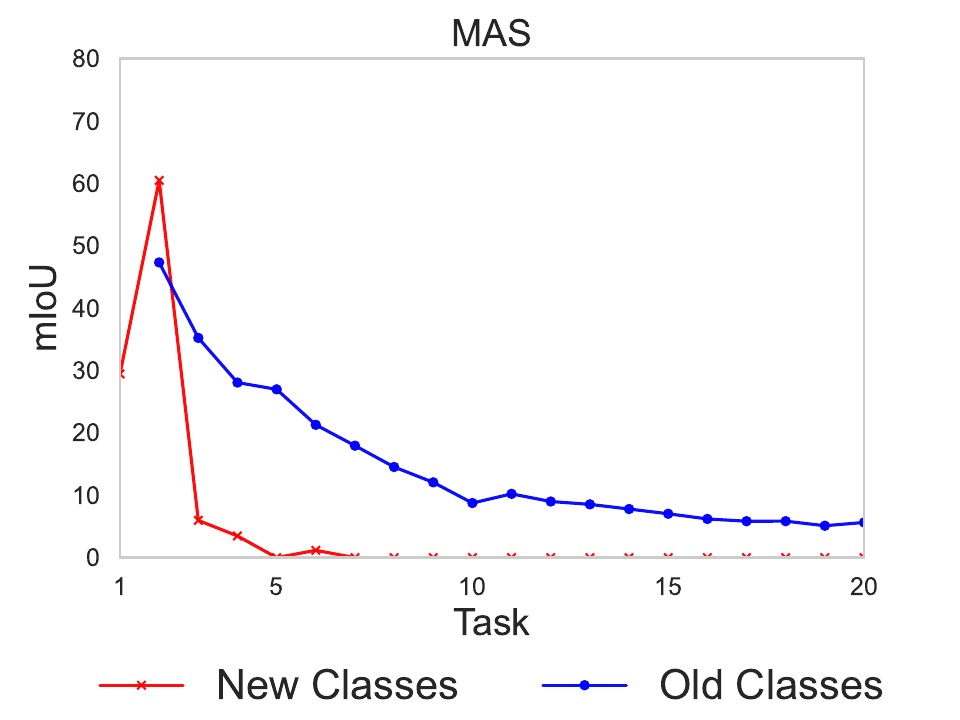}~
\includegraphics[width=0.5\linewidth]{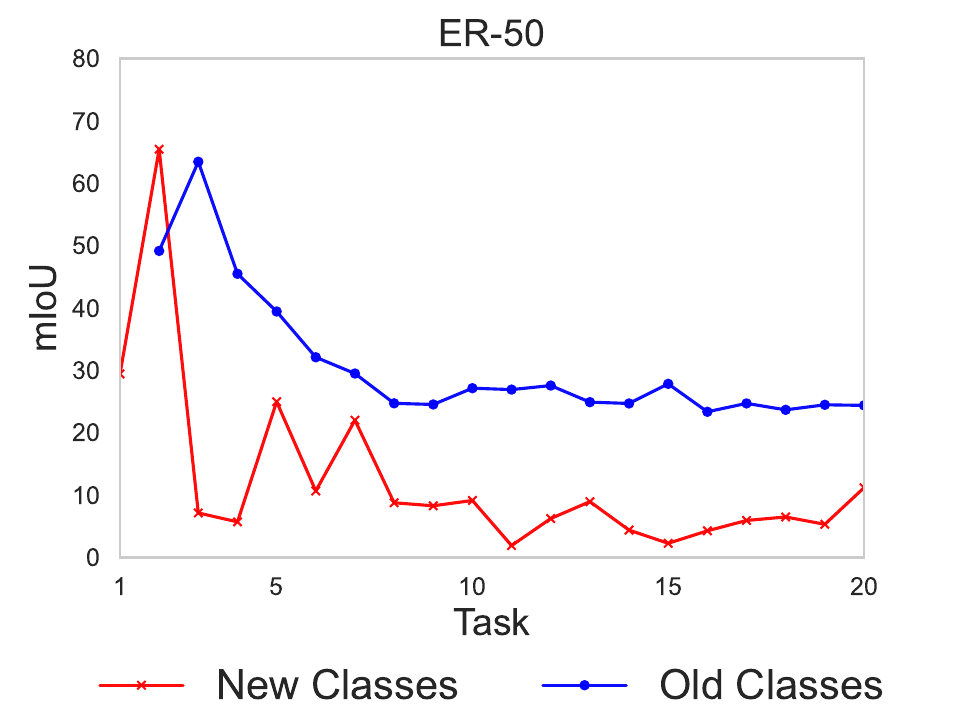}
\vspace{-20px}
    \caption{\color{blue}{Old }\color{black} vs \color{red}{new }\color{black} class performance with MAS~\cite{Aljundi_2018_ECCV} and ER~\cite{chaudhry2019continual}. This provides a clearer understanding of the plasticity and rigidity
of the method. MAS sacrifices plasticity~(learning of new classes) for stability. But even in the best performing ER the learning of the new classes is limited in an online scenario.}
    \label{fig:newold}
\end{figure}
While memory replay methods appear to be at the forefront of continual learning, with discussions extending to the concept of infinite memory~\cite{prabhu2023online},  we would like to point out that this is not solving the fundamental issues of learning continuously and adapting with few data but instead reformulated the problem as another joint training setup.\\
\textbf{How much memory do we actually need?}
As seen Tables~\ref{tab:scannet_miou},\ref{tab:s3dis_miou},\ref{tab:kitti_iou} and Fig.~\ref{fig:total}  the memory size  is a relevant factor, at least up to a certain threshold, especially as the number of tasks increases.
However, in Tab.~\ref{tab:lowmem}  we see that storing just a single example for each class ER-1~(20 examples in total for ScanNet-v2), is comparable to ER-50~(50 examples per class) in the low task regime of $\leq 5$ tasks. \\
\textbf{Can large-scale language models help?} Inspired by the recent advances in open-world segmentation with the use of image-language models like CLIP~\cite{Peng2023OpenScene} we co-embedded our 3D point features with text from the CLIP feature space to alleviate catastrophic forgetting since CLIP features are already easily separated~(Sec.~\ref{sec:clip} for details). Our scenario is much more challenging: we do not use images and the classes for regularization appear separately. But with only 10 points per class for supervision language guidance offers comparable results to the other methods~(Tables~\ref{tab:scannet_miou},\ref{tab:s3dis_miou},\ref{tab:kitti_iou}) that are fully supervised. It can also be used with a very small memory buffer to improve memory replay~(Tab.~\ref{tab:lowmem}).\\
\textbf{Can we learn new classes?} Our setup presents significant challenges. Firstly, it is highly imbalanced, with many classes appearing in only a few scenes and with few points. Secondly, in our online configuration, the model encounters the data only once. To ensure a fair comparison and prevent performance bias towards classes with more instances, we reused scenes containing underrepresented classes. This approach ensures that our model encounters an equal number of scenes for each class, aligning with the second point of our setup, which aims to simulate a challenging but fair online scenario. Mastering new tasks with such limited data proves to be as challenging as mitigating forgetting, even in the best-performing ER scenarios~(Fig.~\ref{fig:newold}, Tab.~\ref{tab:forgets}).\\
\textbf{Task order influence.} The overall performance is impacted by the order the classes appear. Therefore, we conduct our experiments using three distinct class orderings: the original task order from the dataset setup, easy-to-hard, and hard-to-easy. Task difficulty is determined by the number of points of each class (more points$\rightarrow$easier task). \textit{Standard} is close to \textit{difficult} and we in general ward against the \textit{easy} as optimistic. We show mean and variance in Fig.~\ref{fig:errorbar}. \\
\textbf{Decoder architecture influence.}
We empirically find the challenge of OCL-3DSS cannot be solved simply with an advanced Transformer decoder. Details in the Appendix.

\section{Conclusion}

In this work, we expose the issues of the common practices on continual learning methods. We provide initial attempts towards characterizing the difficulty of continual learning even in moderate-sized task sequences ($\geq 5$) and learning sequentially tasks with limited data. We recommend future works to evaluate their proposed continual learning methods on such more challenging closer to real-world setups.

{
    \small
    \bibliographystyle{dpaper2024}
    \bibliography{main,3d}
}

\newpage
\appendix
\clearpage

\twocolumn[{%
\renewcommand\twocolumn[1][]{#1}%
\begin{center}
    \includegraphics[width=\textwidth]{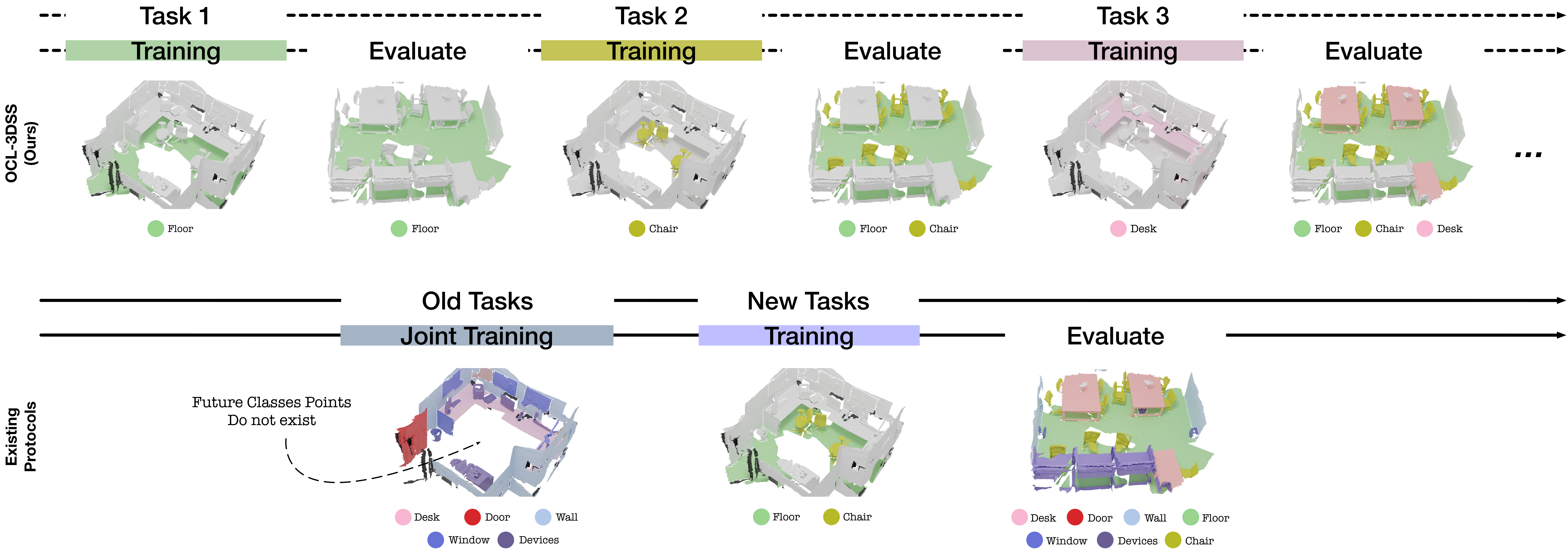}  
    \captionof{figure}{\textbf{An illustration of our continual learning framework.} Our suggested protocol (\textit{top row}) \vs the existing protocols (\textit{bottom row}) 
 (i) learn classes incrementally instead of  jointly training for the majority of classes (ii) evaluates on  long sequences of tasks instead of a two training step approach (iii) contains points belonging to future classes (\textit{floor} and \textit{chair} are missing in $2_{nd}$ row, $1_{st}$ col.)}
    \label{fig:teaser}
\end{center}
}]

\section{Dataset Statistics}
\label{sec:stats}
\begin{figure}
    \centering
    \includegraphics[width=0.45\linewidth]{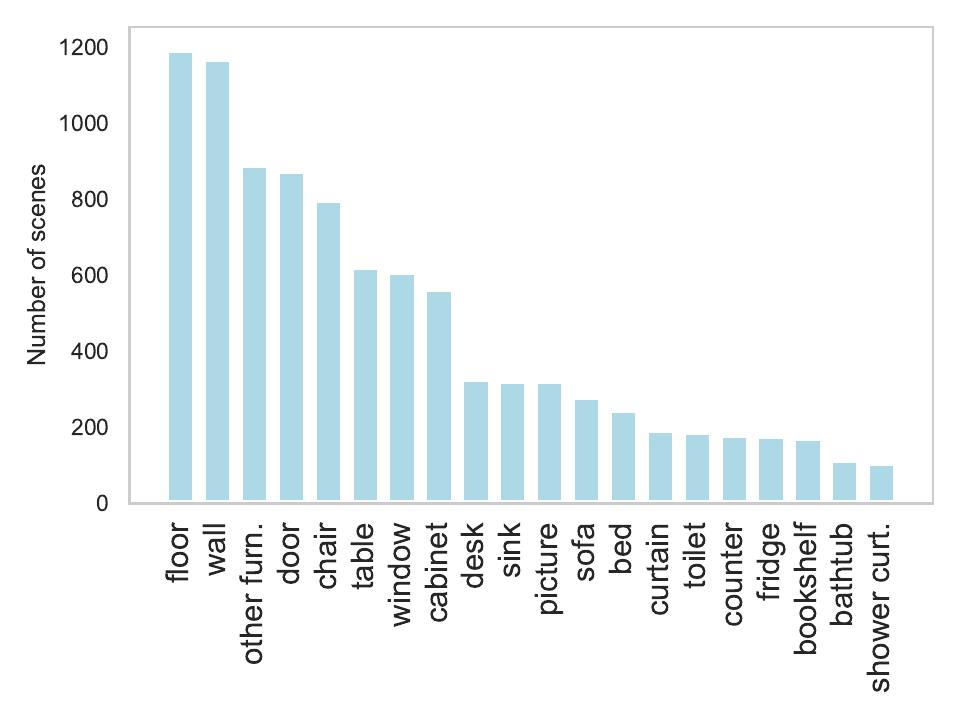}
~
    \includegraphics[width=0.45\linewidth]{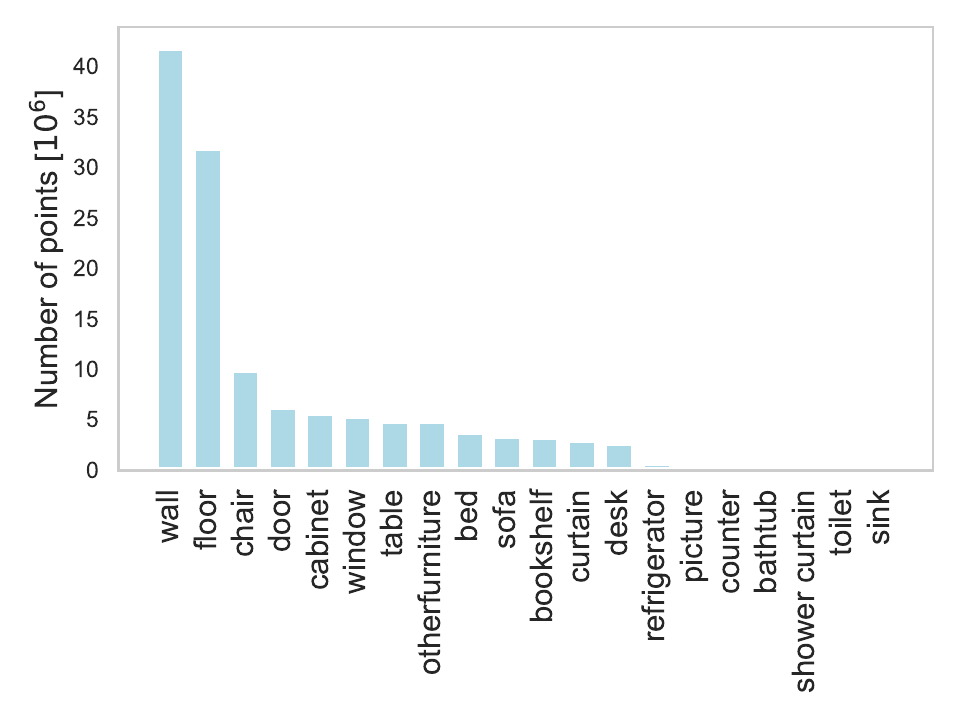}
    \caption{\textbf{ScanNet-v2~\cite{Dai17CVPR}.} Number of scenes a class is present~(\textit{left}) and number of 3D points labeled for each class (\textit{right}). The number of points determines the easyness of a task. The more the points the easier the task. The right figure represents the task ordering of \textit{easy}$\rightarrow$\textit{difficult} scenario described in Sec. \ref{sec:task_diff} in the main paper.}
    \label{fig:scannetv2_stats}

\end{figure}
\begin{figure}
    \centering
    \includegraphics[width=0.45\linewidth]{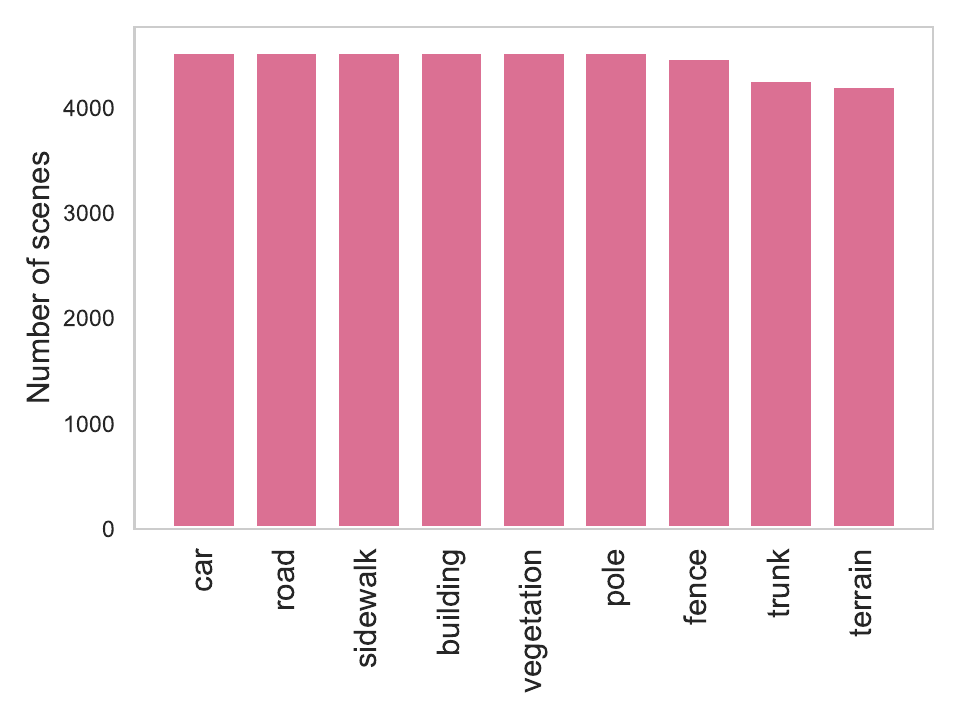}
~
    \includegraphics[width=0.45\linewidth]{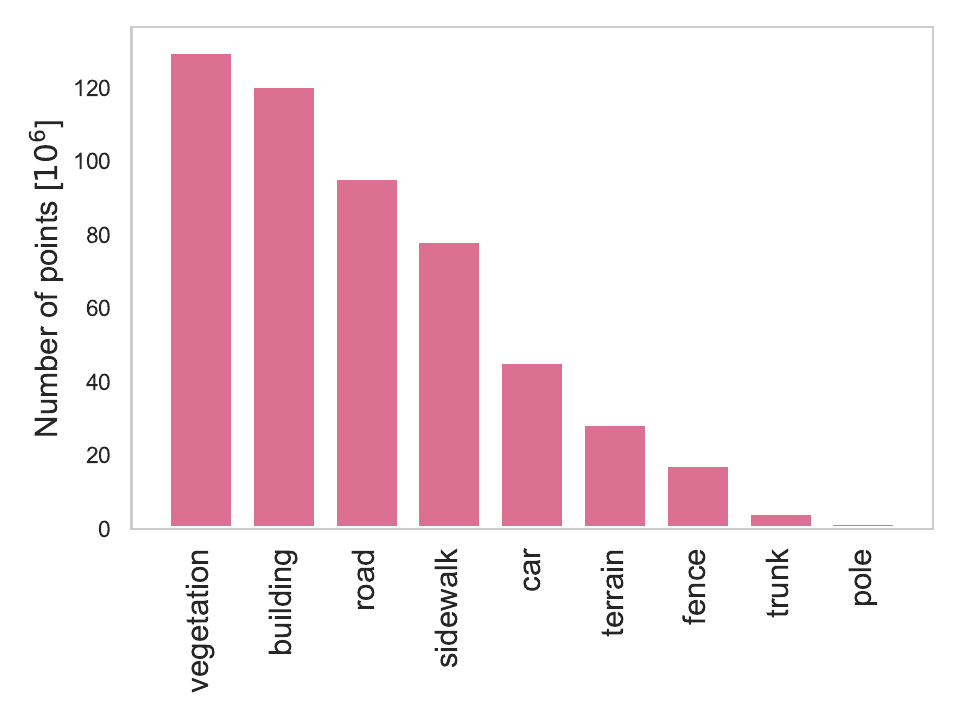}
    \caption{\textbf{SemanticKITTI~\cite{behley2019semantickitti}.} Distribution of class presence across scenes~(\textit{left}) and the number of labeled 3D points for each class (\textit{right}). Our SemanticKITTI setup notably lacks the typical long tail problem~(\textit{left}) encountered in online continual learning, as nearly all 9 classes are present in every scene. When benchmarking OCL, it is crucial to consider not only the point count~(\textit{right}) but also the frequency of scenes featuring each object~(\textit{left}).}
    \label{fig:kitti_stats}

\end{figure}
\begin{figure}
    \centering
    \includegraphics[width=0.45\linewidth]{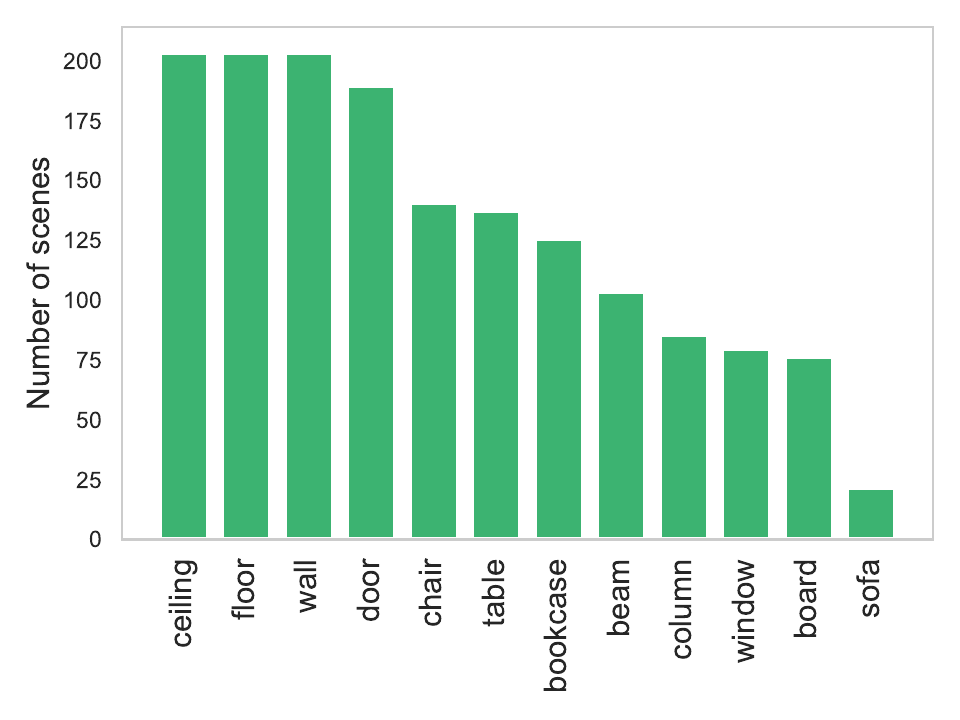}
~
    \includegraphics[width=0.45\linewidth]{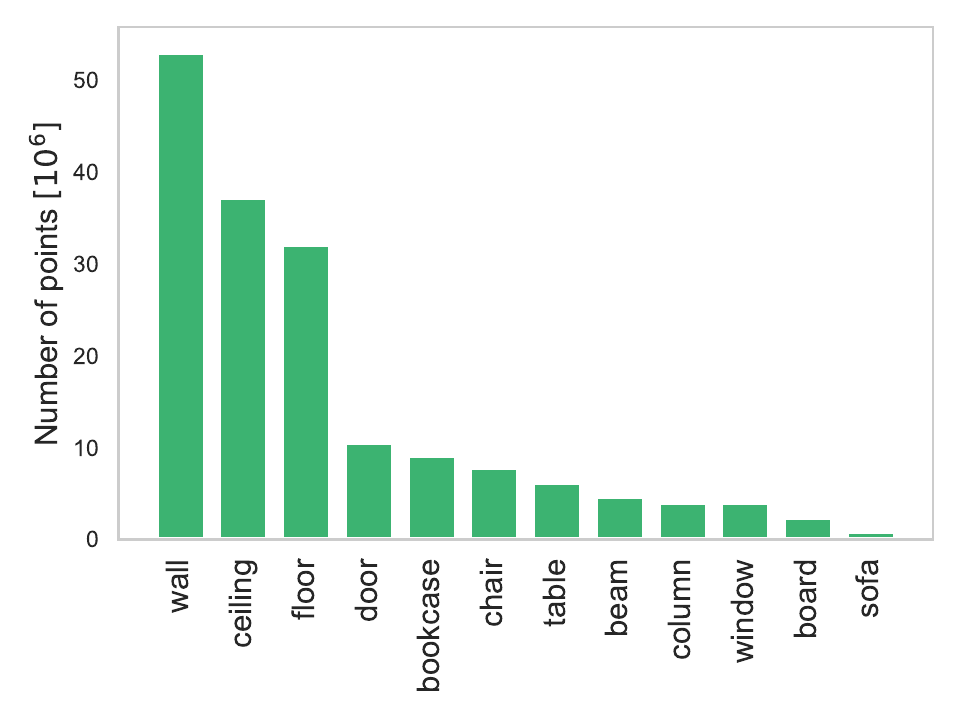}
    \caption{\textbf{S3DIS~\cite{Armeni16CVPR}.} Number of scenes a class is present~(\textit{left}) and number of 3D points labeled for each class (\textit{right}).}
    \label{fig:s3dis_stats}

\end{figure}

\textbf{ScanNet-v2}~\cite{Dai17CVPR} stands out as one of the most extensive indoor 3D datasets, characterized by a long-tail distribution (see Fig.~\ref{fig:scannetv2_stats}, \textit{left}). This distribution, reflecting real-world scenarios, is not only evident in the points per class but also in the frequency of scenes featuring an object class. Real-world occurrences of certain objects are less frequent, making both learning and retaining them challenging. This difficulty is compounded by the fact that these objects don't appear in every scene, even unlabeled in different tasks, rendering methods like pseudo labels~\cite{yang2023geometry} or LwF~\cite{li2017learning} less effective. Due to its alignment with real-world distribution and having the highest number of classes (20), the ScanNet dataset is chosen for \underline{all our ablation studies}.\\
\textbf{SemanticKITTI}~\cite{behley2019semantickitti}, on the other hand, is a large-scale outdoor LiDAR dataset that we employ for training using 4541 scenes from sequence-0. As illustrated in Fig.~\ref{fig:kitti_stats}, SemanticKITTI doesn't exhibit a long-tail problem; most labeled outdoor classes are nearly ubiquitous across scenes. This characteristic mitigates issues related to continual learning, as demonstrated in our evaluation~(see Tab.~\ref{tab:kitti_iou} in the main paper). However, it also highlights that certain datasets, despite their vast size in the number of scenes, may not be ideal benchmarks for evaluating continual learning methods.\\
\textbf{S3DIS}~\cite{Armeni16CVPR} occupies a middle ground between the more diverse datasets of ScanNet and SemanticKITTI, as illustrated in Fig.~\ref{fig:s3dis_stats}. Our training set encompasses a total of 204 scenes for Areas 1-4,6, with Area 5 reserved for evaluation, following standard practices in the field.

\begin{table}

\centering
 \caption{\textbf{OCL-3DSS} on ScanNet-val~\cite{dai2017scannet} after 3, 5, 10 and 20~(final) tasks incorporating an additional \textit{dice loss} alongside cross-entropy. Notably, only ER~\cite{chaudhry2019continual} exhibited improvement.}
    \label{tab:scannet_miou_wdice_appendix}
    \vskip 0.1in
    \resizebox{0.8\columnwidth}{!}{%

    \begin{tabular}{lcccg}
    \toprule

Method &\multicolumn{4}{c}{\textbf{mIoU after \textit{t} tasks}}\\
& \textbf{3} & \textbf{5}  & \textbf{10}  & \textbf{20}\\
\cmidrule{1-5}

\tt{FT(lower bound)} &  24.9 & 3.9 & 2.1 & 0.4\\
\tt{FT  + DiceLoss} & 23.0	 & 1.6 & 0.7 & 0.9\\

\cmidrule{1-5}

\tt{MAS\cite{Aljundi_2018_ECCV}}   & 33.7 & 22.6 & 10.8 & 4.9 \\
\tt{MAS\cite{Aljundi_2018_ECCV}} + DiceLoss& 34.4&19.2  &8.9 
&4.4 \\
\cmidrule{1-5}

\tt{LwF\cite{li2017learning}}  &41.2 & 34.1& 17.4 & 7.8\\
\tt{LwF\cite{li2017learning}} + DiceLoss & 31.5 &25.8  & 16.3 &7.1\\

\cmidrule{1-5}

\tt{ER-200}\cite{chaudhry2019continual}   & 42.2 & 36.6 & 26.8 & 35.7\\

\tt{ER-200\cite{chaudhry2019continual}}+  DiceLoss & \textbf{47.1} & \textbf{41.7} & \textbf{43.8} &\textbf{36.1}\\
\cmidrule{1-5}

\tt{LG-Clip-weak sup.}   & 37.5   & 18.5  
&  10.4 & 3.2\\
\tt{LG-Clip} + DiceLoss &  25.2   & 20.9    &  8.6 & 5.8\\

\cmidrule{1-5}

\tt{JT (upper bound)} &  -& - & -& 66.4 \\
\bottomrule

    \end{tabular}}

\end{table}

\section{Loss Functions}
\label{sec:losses}
In addition to the conventional cross-entropy loss, we assessed our configuration by incorporating the Dice loss~\cite{deng2018learning}~(Tab.~\ref{tab:scannet_miou_wdice_appendix}). While the Dice loss was originally crafted to enhance the definition of boundaries, it also proved beneficial in addressing class imbalances within a standard joint training framework. It demonstrates a modest improvement in scenarios akin to joint learning, such as ER~\cite{chaudhry2019continual}. Nevertheless, it imposes penalties on prevalent, easily identifiable classes, contrasting with other methodologies like LwF~\cite{li2017learning} that rely on them for their performance metrics.

\begin{table}
    \centering
    \caption{\textbf{Online continual learning in 3D semantic segmentation with Transformer decoder~(TD) vs the Convolutional decoder~(CD)} on ScanNet-val~\cite{dai2017scannet} after 3, 5, 10 and 20~(final) tasks.}
\label{tab:scannet_miou_transformer}
\vskip 0.1in
    \resizebox{\columnwidth}{!}{%
\setlength{\tabcolsep}{8pt}

    \begin{tabular}{lrccccg}
    \toprule
Method & \textbf{Mem. Size}& \textbf{Mem./Data}  &\multicolumn{4}{c}{\textbf{mIoU after \textit{t} tasks}}\\
&  $slots\cdot |C|$ &(percent) & \textbf{3} & \textbf{5}  & \textbf{10}  & \textbf{20}\\
\cmidrule{1-7}

\tt{FT+TD(lower bound)} &  - & - & 1.4 & 1.6 & 0.4 & 0.2 \\
\tt{FT+CD(lower bound)} &  - & - & 24.9 & 3.9 & 2.1 & 0.4\\
\cmidrule{1-7}

\tt{MAS+TD\cite{Aljundi_2018_ECCV}} & - &-   &1.4 & 1.6 & 0.8 & 0.4 \\
\tt{MAS+CD\cite{Aljundi_2018_ECCV}} & - &-   & 33.7 & 22.6 & 10.8 & 4.9 \\
  \cmidrule{1-7}
\tt{LwF+TD\cite{li2017learning}}  &- & -& 34.3 & 19.3 & 9.3 & 4.8\\
\tt{LwF+CD\cite{li2017learning}}  &- & -&41.2 & 34.1& 17.4 & 7.8\\
  \cmidrule{1-7}
\tt{ER-50+TD}\cite{chaudhry2019continual} & 50  $\cdot$ 20 & 4.1& \textbf{40.0} & \textbf{35.2} & \textbf{22.9} & 11.5\\
\tt{ER-50+CD}\cite{chaudhry2019continual} & 50  $\cdot$ 20 & 4.1& 29.6& 30.3 & 17.4 & \textbf{17.0}\\

     \tt{ER-200+TD}\cite{chaudhry2019continual} & 200  $\cdot$ 20 & 16.4  & \textbf{50.9} & \textbf{36.6} & \textbf{28.6} & 20.9\\
          \tt{ER-200+CD}\cite{chaudhry2019continual} & 200  $\cdot$ 20 & 16.4  & 42.2 & 36.6 & 26.8 & \textbf{35.7}\\
\tt{ER-400+TD}\cite{chaudhry2019continual} & 400  $\cdot$ 20 & \textbf{32.8}  & \textbf{46.2} & \textbf{40.3} & {33.0} & 28.1 \\
\tt{ER-400+CD}\cite{chaudhry2019continual} & 400  $\cdot$ 20 & 32.8  & {43.6} & {36.4} & \textbf{39.0} & \textbf{42.1}\\
\cmidrule{1-7}

\tt{LG-Clip+TD-weak sup.}  & - & - & 36.1 & 20.8
&  7.7 & 5.2 \\
\tt{LG-Clip+CD-weak sup.}  & - & - & 37.5   & 18.5  
&  10.4 & 3.2\\

\bottomrule

\end{tabular}}

\end{table}

\section{The Importance of the Decoder Architecture}
\label{sec:decoders}

\begin{figure}
    \centering
   \includegraphics[width=0.95\linewidth]{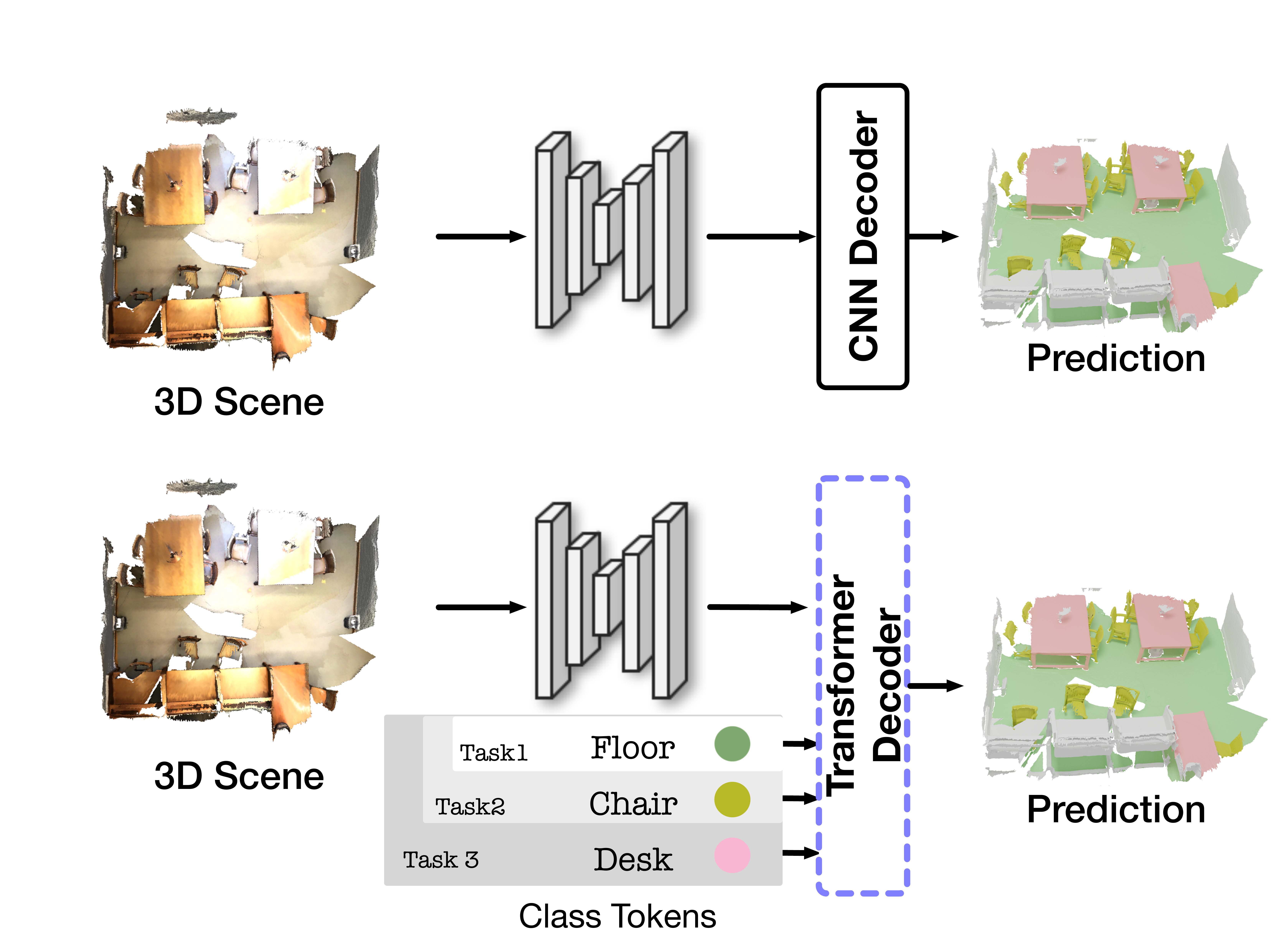}
    \caption{\textbf{ Decoder architectures.} We selected two modern decoder architectures for our study: (i) a conventional convolutional decode~(\textit{top}) and a 
Transformer decoder that has demonstrated state-of-the-art
results in the same field (\textit{bottom}).}
    \label{fig:decoder}

\end{figure}

We selected two modern decoder architectures for our study: (i) a conventional convolutional decoder~\cite{Choy19CVPR} frequently employed in 3D semantic segmentation, and (ii) a Transformer decoder that has demonstrated state-of-the-art results in the same field~(Fig.~\ref{fig:decoder}). Notably, recent efforts~\cite{cermelli2023comformer,shang2023incrementer} have explored leveraging the attention mechanism of Transformers to enhance the stability-plasticity trade-off in continual learning. In the Transformer architecture, we learn a set of class tokens but we dynamically add them to the network one task at a time. Those class tokens interact with encoder features in a Transformer decoder, which will predict the class segmentation mask.

 However, despite the widespread use of Transformer architectures, we illustrate in Tab.~\ref{tab:scannet_miou_transformer} that continual learning challenges persist even when employing such powerful architectures.

The challenges associated with FT are considerably more pronounced in an architecture that demands a substantial amount of data like the Transformer architecture. MAS also encounters significant difficulties in alleviating forgetting in the context of such an architecture. On the other hand, ER exhibits superior performance with the Transformer decoder, however only evident at most up to the $10_{th}$ task. Surprisingly, ER with a transformer decoder appears to face more substantial challenges compared to the convolutional decoder in later tasks, notably in the $20_{th}$ task (last column). This observation underscores \textit{the importance of evaluating a sufficient number of tasks} to make accurate assessments about the effectiveness of different setups in continual learning. Given that the evaluations for 3-5 tasks might not be indicative of performance trends extending to the $20_{th}$ task and beyond.
 
\section{Implementation Details}
\label{sec:details}

\begin{table*}[h!]
\centering
\caption{\textbf{Hyperparameters} for our experiments. LR represents the learning rate and CE the cross-entropy loss. $\lambda_{method}$ is the weight of the regularization loss depending on the method (\eg $\lambda_{mas}$ for the MAS loss explained in Eq.~\ref{eq:mas}). We will publish our code to facilitate reproducibility.}
\label{tab:hyperparameters}
\vskip 0.1in
    \centering
    \resizebox{2\columnwidth}{!}{%
\setlength{\tabcolsep}{2pt}
\begin{tabular}{llcccc}
\toprule
\multirow{2}{*}{Dataset} && \multicolumn{4}{c}{\textbf{Methods}} \\
\cmidrule(lr){3-6}
 & & \tt{MAS}\cite{Aljundi_2018_ECCV} & \tt{LwF}\cite{li2017learning} & \tt{ER}\cite{chaudhry2019continual} & \tt{LG-Clip} \\
\midrule
\multirow{4}{*}{ScanNet~\cite{dai2017scannet}} & LR Encoder & $10^{-4}$ & $5\times10^{-3}$  & $10^{-3}$ & $10^{-3}$  \\
 & LR Decoder & $10^{-5}$  & $10^{-4}$ &$10^{-4}$&$10^{-4}$ \\
& $\lambda_{CE}$ &1  &1  &1  &-  \\
 & $\lambda_{method}$ &  50 & 1  & - &1 \\
\midrule
\multirow{4}{*}{S3DIS~\cite{Armeni16CVPR}} & LR Encoder & $10^{-4}$ & $5\times10^{-3}$ & $10^{-3}$ & $10^{-4}$\\
 & LR Decoder & $10^{-4}$ & $10^{-4}$ & $10^{-4}$ & $10^{-4}$\\
& $\lambda_{CE}$& 1 & 1 & 1 &  - \\
 & $\lambda_{method}$  & 50 &1  & - & 1 \\
\midrule
\multirow{4}{*}{SemanticKITTI~\cite{behley2019semantickitti}} & LR Encoder & $5\times10^{-3}$ & $5\times10^{-3}$ & $10^{-3}$ & $10^{-4}$\\
 & LR Decoder & $10^{-4}$ & $10^{-4}$ & $10^{-4}$ & $10^{-4}$\\
& $\lambda_{CE}$& 1 & 1 & 1 & - \\
 & $\lambda_{method}$  & 1 & 1 &  - & 1\\
\bottomrule
\end{tabular}
}
\end{table*}

\textbf{Model Encoder.} For all datasets and methods, we employ a Res16UNet34A~\cite{Choy19CVPR} as the encoder. The final feature layer has been adjusted to generate features with a dimensionality of 512, deviating from the original 128 in ME~\cite{Choy19CVPR} to align with the CLIP~\cite{radford2021learning} feature dimensionality. Joint Training~(JT) was conducted with the 512 dimensionality, showing no significant performance change compared to the original ME~\cite{Choy19CVPR}.\\
\textbf{Model Decoder.}
The conventional decoder consists of a $1\times1\times1$ convolutional layer with stride 1 that projects encoder features to semantic logits. The Transformer decoder consists of 3 layers with 512 dimensions. Each layer consists of self-attention, cross-attention, and feed-forward network. The attention in each layer has 8 heads.
\\
\textbf{Other Details.}
For the joint training and \cite{yang2023geometry}, the model was trained according to the established pipelines of the original methods.
For the continual learning setup, the network was trained using cross-entropy loss (except in cases where Dice loss was incorporated), employing a stochastic gradient descent optimizer with a mini-batch size of 10. We train the model with the Transformer decoder using AdamW~\cite{loshchilov2017decoupled}. 
Method parameters in detail:
\begin{itemize}
    \item \textbf{ER:} The mini-batch size was reduced to 5, and the mini-batch size retrieved from the memory buffer was also set to 5, resulting in a total of 10, consistent with other methods.
    
    \item \textbf{LwF:} We set the temperature factor \(T = 2\) as per the original paper and other continual learning (CL) papers.
    
    \item \textbf{LG-Clip:} Uses only 10 points per scene for supervision.
    
    \item \textbf{Dice loss:} Weighted with \(\lambda_{\text{dice}} = 50\) and \(\lambda_{\text{CE}} = 1\).
\end{itemize}

For the hyperparameter tuning, we followed~\cite{Ghunaim_2023_CVPR}. For details please see Tab.~\ref{tab:hyperparameters}. 
\textit{We will publish our code upon acceptance to facilitate the reproducibility of our results.}

\subsection{Compared Methods}

In addition to the Joint Training (JT) and Finetuning (FT) baselines outlined in the main paper, we provide a more detailed description of the seminal continual learning works analyzed in our setup.\\
\textbf{Memory Aware Synapses~(MAS).} MAS~\cite{Aljundi_2018_ECCV} incorporates a penalty to regularize the update of model parameters that are important to past tasks. During each training step, MAS optimizes the following loss:
\begin{multline}
\label{eq:mas}
L = \lambda_{CE} \cdot L_{CE}(F(x_k;\theta),\tilde y_t) + \\
\lambda_{mas} \cdot \sum_{i,j} \Omega_{ij}(\theta_{ij}^t-\theta_{ij}^{t-1})^2
    \end{multline}
with $\theta_{ij}^{t-1}$ the parameters of the model for the previous task and $\theta_{ij}^t$ the current model parameters. 
It allows the change of parameters that are not important for previous tasks (low $\Omega_{ij}$) and penalizes the change of important ones (high $\Omega_{ij}$).\\
The parameter importance is computed as:

\begin{equation}
    \Omega_{ij} = \dfrac{1}{N} \sum_{k=1}^N || g_{ij}(x_k)||
\end{equation}

where $g_{ij}(x_k)= \dfrac{ \partial [ l_2^2 (F(x_k;\theta)) ] }{\partial \theta_{ij}}$ the gradient of the learned function with respect to parameter $\theta$ at the data point $x_k$. $N$ is the total number of points at a given task.\\
\textbf{Learning without Forgetting~(LwF).} LwF~\cite{li2017learning} utilizes knowledge distillation from a teacher to a student model to preserve knowledge from past tasks. The teacher model is the model after learning the last task $t-1$, and the student model is the model to be trained on the current task $t$. For a new task $t$ with $(x_k, y_t)$ the input $x_k$ and ground truth $y_t$, LwF computes $F(x_k;\theta^{t-1})$, the output of the previous model $\theta^{t-1}$ for the new data $x_k$. During training, LwF optimizes the following loss:

\begin{multline}
L =   \lambda_{CE} \cdot L_{CE}(F(x_k;\theta),\tilde y_t) + \\
  \lambda_{lwf} \cdot L_{lwf}(F(x_k;\theta^{t-1}),F(x_k;\theta^{t}))
\end{multline}
where $F(x_k;\theta^{t-1}),F(x_k;\theta^{t})$ are the predicted values of the previous and current model using the same $x_k$. The $\lambda$ controls the favoring of the old tasks over the current task. \\
\textbf{Experience Replay~(ER).} ER~\cite{chaudhry2019continual} is a straightforward yet effective replay method. It employs reservoir sampling~\cite{vitter1985random} for updating the memory with new examples and random sampling for retrieving examples from memory. Reservoir sampling ensures that each incoming data point has an equal probability of $(\textit{memory size})/n$ to be stored in the memory buffer. $n$ is the number of data points observed for each task up to the present. In our configuration, all tasks have the same size memory buffer. ER trains the model by combining in the mini-batch the current task data with the data from memory using the cross-entropy loss.

\end{document}